\documentclass[10pt,twocolumn,letterpaper]{article}

\usepackage{cvpr}
\usepackage{times}
\usepackage{epsfig}
\usepackage{graphicx}
\usepackage{amsmath}
\usepackage{amssymb}
\usepackage{lipsum}
\usepackage{etoolbox}

\usepackage[pagebackref=true,breaklinks=true,letterpaper=true,colorlinks,bookmarks=false]{hyperref}
\cvprfinalcopy 
\def\cvprPaperID{1108} 

\DeclareMathOperator*{\argmin}{arg\,min}

\newcommand{\supp}{\textbf{Appendix}~}
\ifcvprfinal\fi
\begin{document}
\AfterEndEnvironment{figure}{\vskip-4ex}

\title{IAN: Combining Generative Adversarial Networks for Imaginative Face Generation}

\author{Abdullah Hamdi, Bernard Ghanem \\
King Abdullah University of Science and Technology (KAUST), Thuwal, Saudi Arabia \\
{\tt\small \{abdullah.hamdi, Bernard.Ghanem\} @kaust.edu.sa}
}

\maketitle

\begin{abstract} 
Generative Adversarial Networks (GANs)  have gained momentum for their ability to model image distributions. They learn to emulate the training set and that enables sampling from that domain and using the knowledge learned for useful applications. Several methods proposed enhancing GANs, including regularizing the loss with some feature matching. We seek to push GANs beyond the data in the training and try to explore unseen territory in the image manifold. We first propose a new regularizer for GAN based on K-nearest neighbor (K-NN) selective feature matching to a target set Y in high-level feature space, during the adversarial training of GAN on the base set X, and we call this novel model K-GAN. We show that minimizing the added term follows from cross-entropy minimization between the distributions of GAN and the set Y. Then, We introduce a cascaded framework for GANs that try to address the task of imagining a new distribution that combines the base set X and target set Y by cascading sampling GANs with translation GANs, and we dub the cascade of such GANs as the Imaginative Adversarial Network (IAN). We conduct an objective and subjective evaluation for different IAN setups in the addressed task and show some useful applications for these IANs, like manifold traversing and creative face generation for characters' design in movies or video games.  


\end{abstract}

\section{Introduction} \label{intro}

\begin{figure}[t!]
    \centering
   \includegraphics[width=\columnwidth]{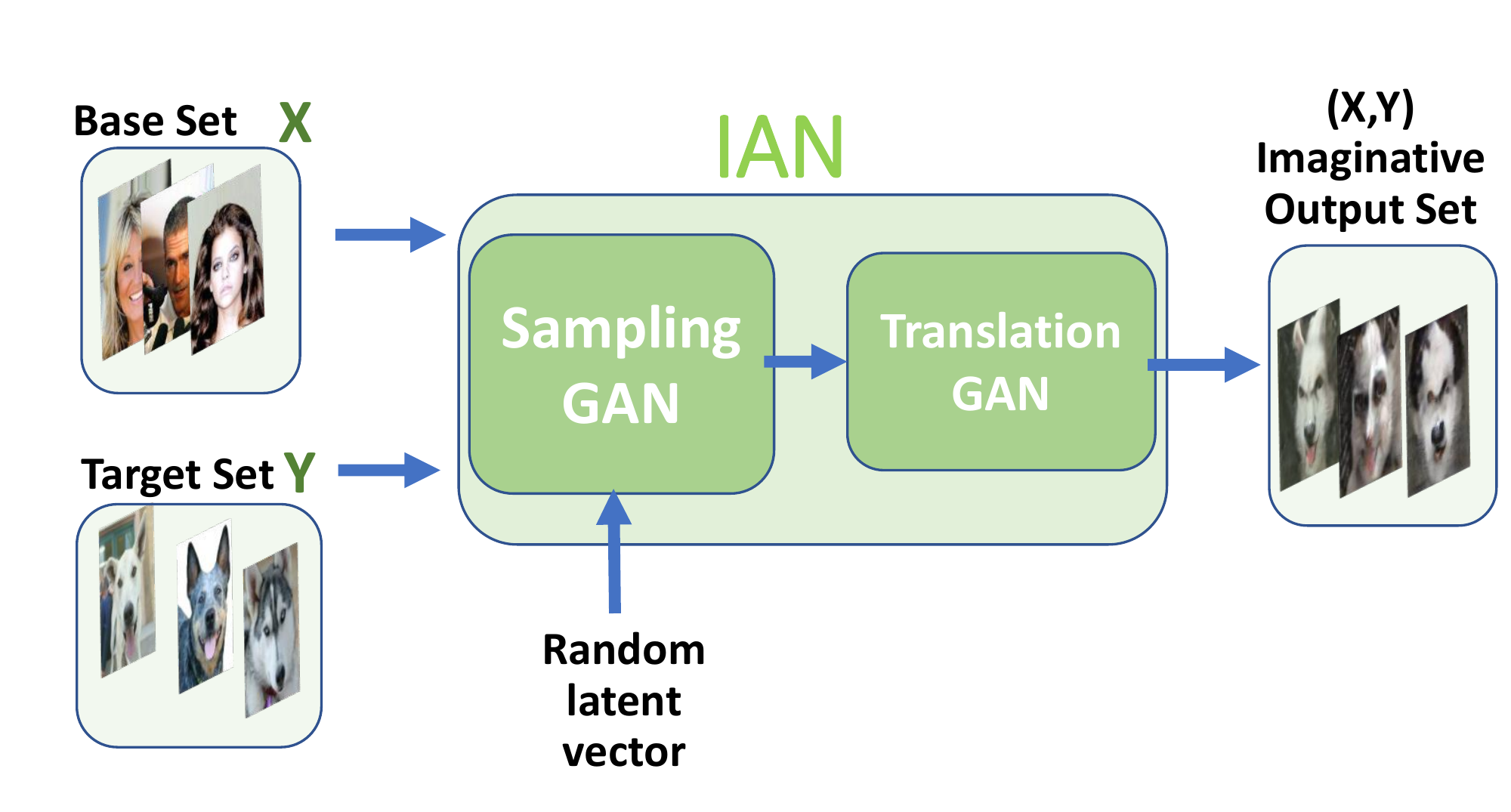}
   \caption{\small We address the task of generating samples from a domain between base set X and target set Y. We do this by introducing cascaded framework of GANs. The first GAN generate samples from learned distribution and the other GAN is conditioned on the output of the first set X to translate to Y. We dub the cascade of GANs for sampling and image translation as the Imaginative Adversarial Network (IAN). The final samples combine characteristics of both sets X and Y}
   \label{IANbest}
\end{figure}

Future AI agents will be interacting with humans and performing more advanced tasks than just classification or detection. Among the qualities of such agents is to push the boundary for human knowledge, using a unique capability that we humans take for granted - namely - \textit{imagination}. Imagination can be roughly defined as the ability to envision what we \textit{have not seen} based on what we \textit{have seen before}. For example, if someone were told to imagine the face of a creature between humans and dogs, they would probably pick a starting creature X ( human or dog) and then think of different alterations for different samples of X to bring them closer to the target creature Y .The flowchart in Fig \ref{IANbest} illustrates this process from which we obtained intuition for our cascaded-GANs framework we call the Imaginative Adversarial Network (IAN).   
\\
\\

Deep neural networks have shown great success in pushing the envelope with regards to discriminative tasks, such as classification and detection \cite{AlexNet,Faster-RCNN}. It is clear that large-scale fully annotated datasets (\eg ImageNet \cite{IMAGENET}) have played a crucial role in this development. However, they are still influenced by biases, which limit their usefulness in the wild \cite{bias}.
The cost of laborious labeling of data (\eg activity localization in videos or pixel-level segmentation) leads us to look for alternative unsupervised ways that can enable further improvements in computer vision tasks. In this context, Generative Adversarial Networks (GANs) have emerged as a promising family of unsupervised learning techniques that have recently shown the ability to model simple natural images, including faces and flowers \cite{GAN}. In the last couple of years, this topic has gained a large amount of attention from the community, and since its inception, many  variations of GAN have been put forward to overcome some of the impediments it faces (\eg instability during training) \cite{WGAN,BEGAN,LaplacianPyrmid,Bigan,inception-GANs2,catGAN,improvGANfeature,zhao2016energy}. Interestingly, GANs have proven useful in several core image processing and computer vision tasks, including image inpainting \cite{inpainting}, style transfer \cite{style-transfer}, super-resolution \cite{img2img,superresolution},manifold traversing \cite{manifold-manipulation}, hand pose estimation \cite{hand-pose}, and face recognition \cite{face-recog}. These success cases indicate that GANs have the potential to learn (in an unsupervised way) important underlying natural image properties that are exploited in turn to synthesize images.

However, one characteristic of the GAN is that it is bounded to the data that it uses for training. In many cases, it might degenerate to simply \emph{memorizing} training data, thus, limiting its ability to discover new ``territory'' in the manifold of natural images. The ability to discover unseen territory is crucial because it can help to bypass the domain adaptation problem of classification \cite{CoGAN}. Although some recent work has been done to maximize innovation in synthesized GAN sample \cite{CAN}, it is limited in scope to the specific domain of art and does not specify the direction of innovation. Our proposed proposed IAN goes beyond this limitation by defining a generic base set X and target set Y. 

Our proposed regularized K-GAN is inspired by recent work on feature matching in adversarial image generation  \cite{PerceptualSim,style-transfer,ActivationMAximization,PlugandPlay,splusu-featurematching,improvGANfeature,progressivegan,spectral-norm}, which  aims to match the generated image to another set of images in high-dimensional feature space. However, Johnson \etal \cite{style-transfer} is optimizing for a single image not a distribution, while Warde-Farley and Bengio \cite{improvGANfeature} matching was bounded to denoising local features of GAN. We utilize more general features of AlexNet \cite{AlexNet} trained on ImageNet\cite{IMAGENET} object classification. Unlike the common practice of modeling base class X and enforcing the generated output to follow certain class label or features of class X \cite{PerceptualSim,ActivationMAximization,PlugandPlay}, we aim to push toward features of target set Y different from X while preserving the GAN modelling of X. Furthermore, most of the methods relied on taking the features of sampled data without validating their quality before matching to them.We propose to follow K-nearest neighbor approach in the high-level feature space in the training of GAN.
Recently, CycleGAN \cite{cycleGAN} showed excellent results in image-to-image translation from source domain X to target domain Y based on cycle consistency regularizer to adversarial loss, but as pointed out in their paper, it fails to model transformation when there are geometrical differences between objects in X and Y. We propose to utilize CycleGAN as the translation GAN in our IAN framework combined with different sampling GAN variations ( including our novel K-GAN ) to achieve the goal of modeling new meaningful distribution between X and Y.
\\\noindent \textbf{Contributions.} \textbf{(1)} We propose a new regularizer for GAN that uses K-nearest neighbor (K-NN) selective feature matching to target set Y in high-level feature space during the adversarial training of any GAN, and we call this novel model K-GAN. We show that minimizing the added term follows from cross entropy minimization between Y and the GAN distribution . \textbf{(2)} We present a cascade framework for GAN to push its output away from the base domain X to target Y. We call this setup the Imaginative Adversarial Network (IAN).  We conduct objective and subjective evaluation for different IAN setups in the addressed task and show some useful applications for these IANs , like multi-domain manifold traversing.   

\linespread{0.99}

\section{Related Work}
\subsection{Generative Adversarial Network}
Generative Adversarial Networks (GANs) \cite{GAN} are generative models consisting of a Discriminator $\mathbf{D}_{X}$ and a Generator $\mathbf{G}$ that are adversarially trained in a similar manner as a minimax game. The discriminator tries to determine if a given image is real (from training) or fake (generated by $\mathbf{G}$). The Generator on the other hand is trying to change the generated samples to better fool $\mathbf{D}_{X}$. This can be formulated as the following minimax game on the loss function $\mathbf{\mathit{L}}_{GAN}(\mathbf{G},\mathbf{D}_{X},\mathbf{P}_{X})$: 
\begin{equation} 
\begin{aligned}
&
\min_{\mathbf{G}} \max_{\mathbf{D}}~~\mathbf{\mathit{L}}_{\text{GAN}}(\mathbf{G},\mathbf{D}_{X},\mathbf{P}_{X})=  \\
& \mathbb{E}_{\mathbf{x}\sim p_{x}(\mathbf{x})} [\log \mathbf{D}(x)] +  \mathbb{E}_{\mathbf{z}\sim p_{\mathbf{z}}(\mathbf{z})} [\log (1- \mathbf{D}(\mathbf{G}(z)))]
\end{aligned}
\label{eq:GAN}
\end{equation}
\noindent where $\mathbf{P}_{X}$ is the distribution of images in domain X, $ \mathbf{z} \in \mathbb{R}^{d}$ is the latent uniformly random vector.
Zhao \etal introduces energy-based loss for GANs \cite{EBGAN}, and recent GANs uses the Wasserstein distance instead of K-L divergence as objective \cite{WGAN,BEGAN}. Both EBGAN \cite{EBGAN} and BEGNA \cite{BEGAN} uses auto-encoder Discriminator network with its loss defined as L2 pixel difference between input and output. 
\subsection{Feature Matching (FM)}
The method of Warde-Farley and Bengio \cite{improvGANfeature} uses Denoising Auto Encoder (DAE) feature matching to enhance the training of the generator and to increase the diversity of the samples. The work of Dosovitskiy A. and Brox T. \cite{PerceptualSim} shows that a conditioning network $\mathbf{C}$ (also called Comparitor) can be incorporated in the GAN loss by adding a discrepancy loss in deep feature space, which measures the difference between the generated image and  same class images in this feature space to increase Perceptual similarity. A similar technique is used for style transfer and super resolution by Johnson \etal \cite{style-transfer}. They try to match covariance matrix of deep features of some image (extracted from  VGGNet \cite{VGG})and match that to some target domain. Shrivastava \etal  \cite{splusu-featurematching} uses GAN loss to improve the quality of synthetic images and used the deep features of a network trained on classifying eye gaze to maintain the key elements of the synthetic data (\eg the direction of gaze). Plug\&Play \cite{PlugandPlay} uses DAE to model the FC7 and Conv5 layers of  $\mathbf{C}$ network and shows that it can improve the quality of the generator network under some conditions.
\subsection{Guiding the Generator to New Distributions }
As pointed out by Elgammal \etal \cite{CAN}, GANs by nature emulate but do not innovate, hence their work introduces the Creative Adversarial Network (CAN) to tackle this issue. They add a loss to the traditional GAN loss to encourage the generator to increase its entropy so as to fool the discriminator, leading the generator to produce more innovative samples. We tackle the same problem but through a different approach. As compared to \cite{CAN}, we define a target set Y with distribution $\mathbf{P}_{Y}$ (different from the data distribution $\mathbf{P}_{X}$) that guides the generator to new territory. In this context, the work of LIU and Tuzel \cite{CoGAN,new-coupled-gan} introduces CoGAN to model the joint distribution by coupling two GANs in parallel with shared network weights. This aids domain adaptation when simple distributions are available (\eg MNIST \cite{MNIST}). However, it does not work well for more complex domains as compared to CycleGAN \cite{cycleGAN}. Bang \etal try to translate images of one class to another class that is has similar attributes by learning joint features and then matching the mean features of the two classes \cite{resemblegan}
Unlike image manipulation \cite{face-editing}, which manipulates individual image at a time, we manipulate the entire distribution of $\mathbf{G}$ from which we can sample.
\subsection{Cycle-Consistent Adversarial Networks}
Recently, CyclGAN \cite{cycleGAN} has shown compelling results in image translation by imposing cycle consistency between two generators (each takes images as input). It produces a sharp translation from one domain X to another domain Y with no paired training. CycleGAN has the following loss on the two distributions $\mathbf{P}_{X}$ and $\mathbf{P}_{Y}$:
\begin{equation}
\begin{aligned}
&\mathbf{\mathit{L}}_{\text{CycGAN}}(\mathbf{A},\mathbf{B},\mathbf{D}_{X},\mathbf{D}_{Y}) =\\ &\mathbf{\mathit{L}}_{\text{GAN}}(\mathbf{A},\mathbf{D}_{X},\mathbf{P}_{X})  +
\mathbf{\mathit{L}}_{\text{GAN}}(\mathbf{B},\mathbf{D}_{Y},\mathbf{P}_{Y}) +
\lambda ~ \mathbf{\mathit{L}}_{\text{co}}(\mathbf{A},\mathbf{B})    
\end{aligned}
\label{eq:CycleGAN}
\end{equation}
where $\mathbf{A}, \mathbf{B} $ are the two generators, $\mathbf{D}_{X}$ and $\mathbf{D}_{Y}$ are the discriminators for domains $X$ and  $Y$ respectively. $\mathbf{\mathit{L}}_{co}(\mathbf{A},\mathbf{B}) $ is the cycle consistency loss of the CycleGAN that encourages that $ \mathbf{B}(\mathbf{A}(\mathbf{x})) = \mathbf{x}$ and is defined as:
\begin{equation}
\begin{aligned} \label{eq:CycleGAN-cons}
\mathbf{\mathit{L}}_{co}(\mathbf{A},\mathbf{B}) &=  
\mathbb{E}_{\mathbf{x}\sim p_{\mathbf{x}}(\mathbf{x})} [ \| \mathbf{B}(\mathbf{A}(\mathbf{x}))) - \mathbf{x}  \|_{1}] \\ &
+\mathbb{E}_{\mathbf{y}\sim p_{\mathbf{y}}(\mathbf{y})} [\| \mathbf{A}(\mathbf{B}(\mathbf{y}))) - \mathbf{y}  \|_{1}]]
\end{aligned}
\end{equation}
where $\lambda$ is  controls the consistency regularizer.
Although CycleGAN shows compelling results when the images translation occurs in texture or color, it shows less success when the translation requires geometric changes. This is due to the limited features modeling CyclGAN performs.A more recent work by Huang \etal utilized cycle consistency in feature space by separating content and style  to produce varying translations \cite{multigan} .The work of Choi \etal utilizes CycleGAN with multiple classes \cite{stargan}. Gokaslan \etal improve shape deformations for CycleGAN\cite{ganiphorm}.
We address CycleGAN limitation in our proposed method by enforcing change in the object's deep features, which tend to go beyond mere texture representation. We do this by introducing K-nearest neighbor feature matching (KNN-FM), which encourages the generated sample to be close in deep feature space to its NN in the target set. Interestingly, we also see that this type of matching can also be viewed mathematically as a way to reduce the cross-entropy between the target distribution $\mathbf{P}_{Y}$ and the distribution of the generated images. 
\begin{figure*}[t!]
\begin{center}
\includegraphics[width = \linewidth]{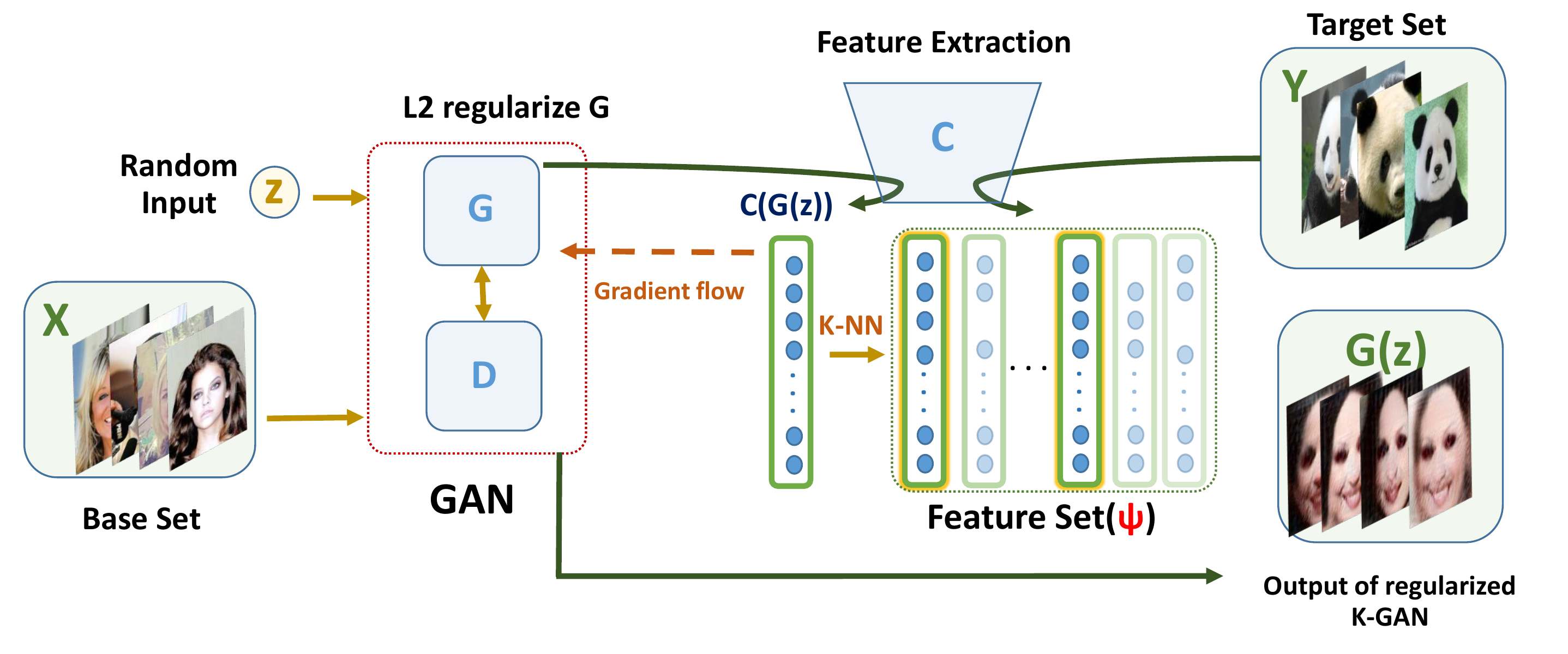}
\end{center}
     \caption{\small K-GAN as a flavoured sampling GAN : We propose K-NN regularizer for GAN that models a set X to push it further toward target set Y. In red frame left is the standard GAN of Generator G, Discriminator D. We add regularizer (right) with deep features extracted by conditioning network C on target Y and output of G. The target of regression in deep feature space is selected based on K-NN search in the Features set of Y. Samples of K-GAN maintain X main appearance but obtain some Y properties (\eg the pandas eyes and nose above). The output of regularized G can then be passed to image translation network e.g. CycleGAN to obtain a full IAN .}
\label{fig:pipeline}
\end{figure*}
\section{Methodology}
We propose Imaginative Adversarial Network (IAN), a two-stage cascade imaginative model that aims to generate samples that resemble base set X and target set Y. First, we sample from a GAN by a latent vector $\mathbf{z}$, then the second stage is a translation GAN, that takes the samples from the first stage and translate them to a different domain.In the first stage (sampling stage), we use the classical DCGAN\cite{DCGAN} and recent BEGAN\cite{BEGAN}, along with their K-NN regularized versions . We use the recent CycleGAN \cite{cycleGAN} as a second stage with both stages trained on the same X and Y sets and cascaded during the sampling to obtain an output that both have key properties of X and Y as illustrated in the flowchart in Fig \ref{fig:pipeline}. For start , we develop the K-NN regularized GAN in the following subsection. 
\subsection{K-GAN} \label{KGANsection}

K-GAN adds to the traditional GAN loss in Eq (\ref{eq:GAN}) a term that encourages the generator distribution $\mathbf{P}_{g}$ towards a target distribution $\mathbf{P}_{Y}$ through cross entropy minimization. The similarity of $\mathbf{P}_{g}$ to $\mathbf{P}_{X}$ is also encouraged through the traditional GAN loss. The K-GAN loss is defined as follows:
\begin{equation}
\begin{aligned} \label{eq:KGAN}
&\mathbf{\mathit{L}}_{\text{KGAN}}(\mathbf{G},\mathbf{D}_{X},\mathbf{P}_{X},\mathbf{P}_{Y})  = \\ &\mathbf{\mathit{L}}_{\text{GAN}}(\mathbf{G},\mathbf{D}_{X},\mathbf{P}_{X})   +
 \mu ~ \mathbb{E}_{\mathbf{z}\sim p_{\mathbf{z}}(\mathbf{z})} [H(\mathbf{G}(z) , \mathbf{P}_{Y})]
\end{aligned}
\end{equation}

where the added term is the cross entropy between the image distribution of our generator $\mathbf{P}_{g}$ and the image distribution of the target $\mathbf{P}_{Y}$, and $\mu$ is a regularizing hyperparameter. We replace the cross entropy $H(\mathbf{G}(z),\mathbf{P}_{Y})$ with the following K-NN regularizer:
\begin{equation} \label{eq:xentmin}
\begin{aligned}
  &\mathbf{\mathit{L}}_{\text{KNN}}(\mathbf{G},\mathbf{P}_{Y}) = \\
  &\mathbb{E}_{\mathbf{z}\sim p_{\mathbf{z}}(\mathbf{z})} \sum_{n=1}^{k} \| \mathbf{C}(\mathbf{G}(z)) - \mathbf{r}_{n}\left(\mathbf{C}\left(\mathbf{G}\left(z \right)\right) \right)  \|_{2}^{2}
\end{aligned}
\end{equation}
This is the distance between a generated sample from $\mathbf{G}$ by a latent vector $\mathbf{z}$ and its K-NN samples of the target distribution $\mathbf{P}_{Y}$ from the total $M$ samples available from $\mathbf{P}_{Y}$. This distance is not in pixels, but in high-level feature space defined by the Comparitor network $\mathbf{C}(\mathbf{b}) \in \mathbb{R}^{s}$ which takes an image $\mathbf{b} \in \mathbb{R}^{h \times w}$ as input and outputs feature vector (in our case it was FC7 and FC6 of AlexNet \cite{AlexNet} trained on ImageNet \cite{IMAGENET}). Refer to Fig \ref{fig:pipeline} for visualization. $\mathbf{r}_{n}(\mathbf{c})$ is the K'$^{\text{th}}$ NN function, a parametric (with parameter $n$) order selection function that selects an element $\mathbf{y}_{i}$ from a set of $M-n+1$ elements based on how close it is to input vector 
$\mathbf{c}$. It can be described by the following:
\begin{equation}\label{eq:KNN}
    \mathbf{r}_{n}(\mathbf{c}) = \argmin_{\mathbf{y}_{i} \in \mathbb{\psi}_n} \| \mathbf{c} - \mathbf{y}_{i} \|_{2},
\end{equation}
where $\mathbb{\psi}_{n} $ is the set of deep features of $M-n+1$ images representing the target distribution $\mathbf{P}_{Y}$. The optimization in (\ref{eq:KNN}) is discrete and the resulting function $\mathbf{r}_{k}(\mathbf{c})$ is complicated and indifferentiable but fixing it to current estimate of $\mathbf{G}$ solve the problem. See Sec.\ref{training} for training details.

\noindent To show how we get Eq.(\ref{eq:xentmin}), first we look at $\mathbf{P}_{Y}$. Since acquiring a full description of $\mathbf{P}_{Y}$ in deep feature space is infeasible in most cases, we settle to approximate it using $M$ target samples. We can get a proxy distribution of $\mathbf{P}_{Y}$ by using Kernel Density Estimation (KDE) on the deep features extracted by deep network $\mathbf{C}(\mathbf{x})$ from those $M$ samples in $\mathbf{P}_{Y}$. By picking Gaussian kernel for the KDE, the proxy distribution estimator $\hat{\mathbf{P}}_{Y,\text{proxy}}$ is defined by a Bayesian non-parametric way at any point $\mathbf{C}(\mathbf{b}) \in \mathbb{R}^{s}$ for $\mathbf{b \in \mathbb{R}^{h \times w}}$ as follows:
\begin{equation} \label{eq:proxy}
\hat{\mathbf{P}}_{Y,\text{proxy}}(\mathbf{b}) = \frac{1}{M\sigma \sqrt{2\pi}}\sum_{i=1}^{M} \exp\left(-\frac{\|\mathbf{C}(\mathbf{b}) - \mathbf{C}(\mathbf{y}_{i})\|_2^2}{\sigma^2}\right)  
\end{equation}
\noindent $\mathbf{y}_{i}$ is the $i^{\text{th}}$ sample describing $\mathbf{P}_{Y}$ and $\sigma^{2}$ is the Gaussian kernel variance. 
For far images from the $M$ samples of  $\hat{\mathbf{P}}_{Y,proxy}$ the probability becomes exponentially negligible. Hence, we we can investigate the effect of a subset of K points (out of the M points $\mathbf{C} (\mathbf{y}_{i})$) that are closest to $\mathbf{C}(\mathbf{b})$ to compute (\ref{eq:proxy}). We replace $M$ by $K$ in (\ref{eq:proxy}), and then use Jansen inequality on that modified KDE to get a bound on the expression. By replacing $\mathbf{b}$ by $\mathbf{G}(\mathbf{z})$ and taking logarithm of expectation of both sides, we obtain upper bound on the cross entropy $\mathbb{E}_{\mathbf{z}\sim p_{\mathbf{z}}(\mathbf{z})} [H(\mathbf{G}(z), \mathbf{P}_{Y})]$. The bound found is the one in (\ref{eq:xentmin}), and by minimizing that bound we minimize the cross entropy in Eq.(\ref{eq:KGAN}). For a detailed derivation, please see the \supp.

\noindent The K-GAN objective is therefore defined as follows: 
\begin{equation}
\begin{aligned} 
 \min_{\mathbf{G}} \max_{\mathbf{D}}& ~ ~\mathbf{\mathit{L}}_{\text{GAN}}(\mathbf{G},\mathbf{D}_{X},\mathbf{P}_{X})  + \mu  ~ \mathbf{\mathit{L}}_{\text{KNN}}(\mathbf{G}, \mathbf{P}_{Y}),
\label{eq:final-KGAN}
\end{aligned}
\end{equation}
\noindent where the two terms follow from (\ref{eq:GAN}) and (\ref{eq:xentmin}) respectively.

\noindent The goal of employing KNN-FM in our loss is to utilize the distribution obtained by deep features of some set Y to guide the generation of the GAN. If GAN generated samples are close to all samples of target set Y in deep features, we become more confident about the "objectiveness" of our samples. However, measuring the distance to all samples features in a set is computationally expensive, especially in an iterative optimization like what is in GAN training. Hence, following the neighbors to guide the generation seems a viable alternative to the expensive global judgment of all samples.
\\ \indent Using K-features rather than a single feature at each training iteration will further improve the robustness. This is due to the fact that features at layers (\ie FC6 and FC7) of AlexNet are prone to small imperceptible noise. To demonstrate this, we add uniform noise with a maximum pixel corruption of $4\%$ to 10000 randomly selected images from the ImageNet validation set. This resulted in a relative change in intensity values as high as 20\%.
This shows that the high-level feature space is prone to noise and depending on that solely can lead us astray from the objective we seek. That is why many previous works \cite{PlugandPlay,improvGANfeature} adapted the DAE in their architectures to mitigate the volatility of deep features. So, using the K-nearest neighbor goes along the same direction with smoothing parameter $\mathbf{K}$.
\subsection{IAN Cascade} \label{cascasde}
\fontdimen4\font=1pt As described in Sec.\ref{intro}, an IAN model consists of a sampling GAN and a translation GAN to generate samples that capture the properties of two sets X and Y. We utilize our K-GAN model developed in Sec.\ref{KGANsection},along with other Vanilla GANs, and cascade them with the recently developed cycleGAN \cite{cycleGAN} after training each separately with some fine-tuning (see Sec.\ref{training} for training details). The goal is to utilize the ability of CycleGAN to transfer appearance for our sake to generate natural looking samples that capture key properties and appearance of both X and Y. By using K-GAN as input to the CycleGAN, the later gains these advantages:
\\ \indent \textbf{Sampling.} $\mathbf{z}$ vector sampling is a property of GAN to sample from the modeled distribution. However, CycleGAN is conditioned on input images.Hence, we use the sampling in K-GAN and transform the output with CycleGAN. While this can be done by any GAN, we show in Sec.\ref{results-section} how the combination of K-GAN+CycleGAN outperforms others in many metrics.\\ 
\indent \textbf{Object geometry.} As pointed out by its authors, CycleGAN is limited to the appearance transfer when it comes to cross-class translation (\eg transforming cat to dog) \cite{cycleGAN}. We try to tackle this by enforcing feature matching to the target set, which results in geometric change that is appended by the appearance transfer of CycleGAN.
\indent \textbf{Regularization.} While the CycleGAN is trained on natural images, it is used here to translate synthetic images produced by other GANs. This poses the risk of amplifying the imperfections produced by the first GAN. We show in Sec.\ref{evaluation} that regularizing the GAN (as in K-GAN) limits the possible outputs which helps in mitigating this effect of error amplification when CycleGAN is fine tuned to the regularized samples. The diversity of the final samples is achieved by the cascade effect with CycleGAN.

\section{Experiments} \label{experemtn}
\subsection{Implementation Details} \label{Implementation}
For the K-GAN part, we use the basic network implemented in DCGAN \cite{DCGAN} and the more advanced architecture in BEGAN\cite{BEGAN} . The KNN-FM term makes use of deep features extracted from the FC6 and FC7 layers of AlexNet \cite{AlexNet} pre-trained on ImageNet\cite{IMAGENET}. 
AlexNet was favored over deeper networks because at each iteration of training K-GAN, feature extraction, multiple forward passes,  and backpropagations are required for updating $\mathbf{G}$. We used FC7 because it semantic information about the object and used usually for discriminative tasks \cite{FC7}. We used FC6 because it helped in transferring the key properties of the object along with FC7. We use the same CycleGAN network as proposed in \cite{cycleGAN}, in which they used the PatchGAN discriminator implemented in \cite{img2img} and U-Net Generators. The image output samples all have 227$\times$227 pixel resolution.
\subsection{Training }\label{training}
\textbf{Training Data}\\
We used CelebA \cite{CelebA} as our base set X and we used 10 different animal classes from ImageNet \cite{IMAGENET} as our target set Y. The Y dataset was manually filtered to be suitable for the task of imagining humanoid animals by ensuring that the face of the animal appears in the image. Each of the Y target classes contains 100 to 500 images. The Y classes are gorillas, chow chow dog, Egyptian cat, koala ,lion, Norwegian-elkhound dog, panda, Samoyed-dog, spider, and tiger. They were used because of their distinctive features, the abundance of facial data, and range of difficulty. We call this filtered dataset \textit{Zoo-Faces}, and it will be made publicly available with the code. Refer to the \supp for some qualitative examples from this dataset. We tried to use ImageNet \cite{IMAGENET} without filtering, and that did not help to achieve the objective, and therefore we had to use filtered data collected from the internet. See Sec.\ref{ablation} for more details. Before training, we take the Zoo-Faces dataeset and extract the FC7 and FC6 features by the Comparitor $\mathbf{C}$ (AlexNet) to obtain 20 different feature matrices of FC7 and FC6 for the 10 classes. These feature sets are then passed to the K-GAN to be used in training.
\\
\\
\begin{figure}[t]
\includegraphics[scale=0.2]{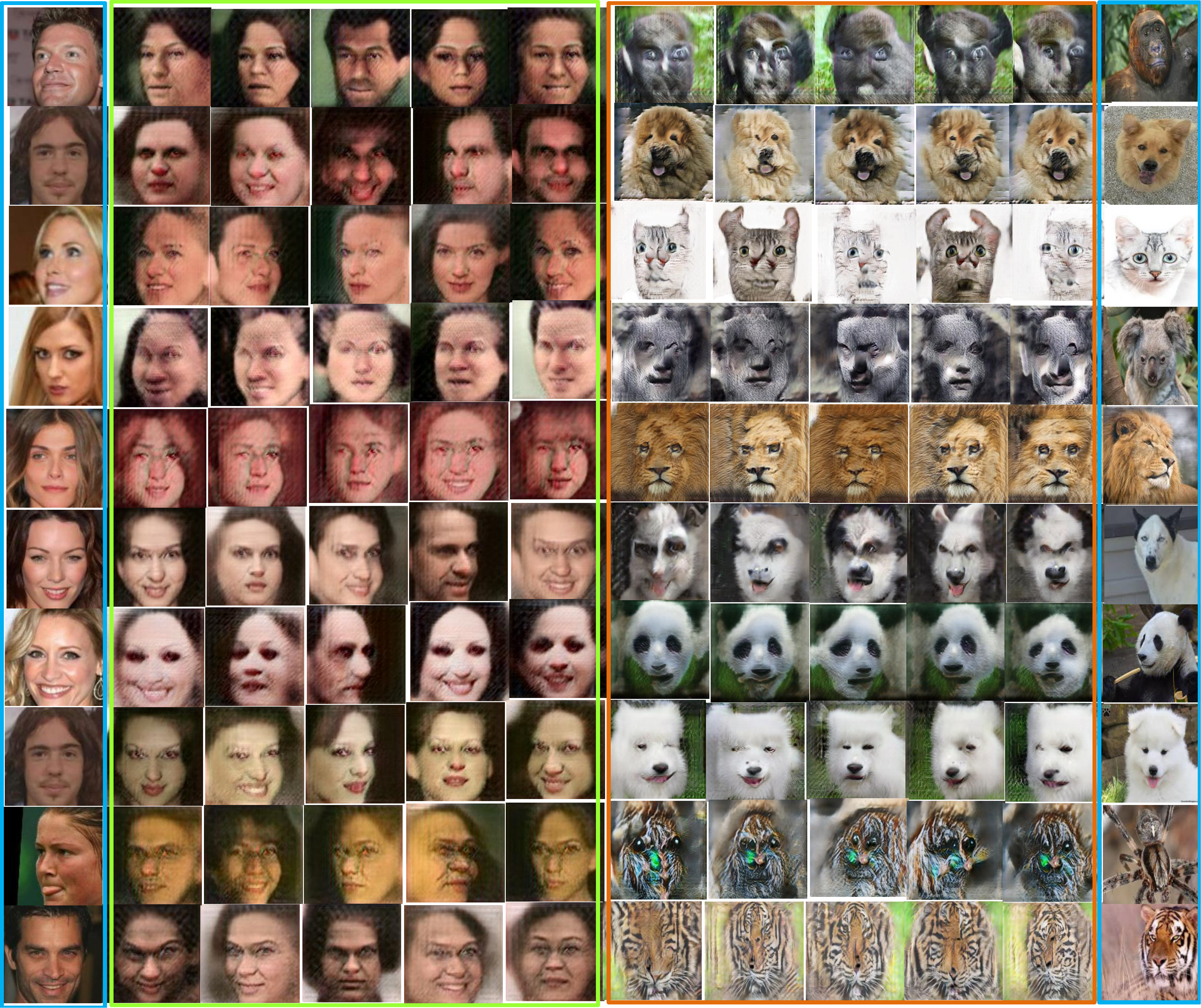}
  \caption{\small Samples from K-BEGAN and its corresponding IAN: green-framed columns are K-BEGAN samples, orange-framed columns are full-IAN samples, each row is one X-Y IAN model where X:faces and Y is a class in zoo-faces. The far left and far right blue framed columns are the closest training examples to the neighboring K-GAN and IAN samples respectively.}
\label{qualititative}
\end{figure}
\textbf{Training Procedure}\\
Since we are running a gradient descent based algorithm (Adam optimizer \cite{adam}), by fixing the argument of $\mathbf{r}_{k}(\mathbf{.})$ in Eq.(\ref{eq:KNN}) to the current estimate of the generator  $\mathbf{G}_{t}$ at time step t, the term $\mathbf{r}_{n}(\mathbf{C}(\mathbf{G}_t(\mathbf{z})))$ is just a constant $\mathbf{d}_{n,t}$ at that iteration. For step t and parameter $n=K$, we use the ball-tree algorithm \cite{balltree} to search for K-NN features in the previously obtained set $\mathbb{\psi}_{n}$ that are close to the current sample from $\mathbf{G}_t(\mathbf{z})$. The term in Eq.(\ref{eq:xentmin}) becomes just an $\ell_2$ regularizer: 
\begin{equation} \label{eq:xentmin_current}
  \mathbb{E}_{\mathbf{z}\sim p_{\mathbf{z}}(\mathbf{z})} \sum_{n=1}^{k} \| \mathbf{C}(\mathbf{G}(z)) - \mathbf{d}_{n,t}  \|_{2}^{2}
\end{equation}
For practical reasons, we minimize the distance between $\mathbf{C}(\mathbf{G}(z))$ and the average of the $K$ neighbors instead of the distances to all of them independently. This is justified and can be seen by expanding the norms and looking argument minimizer to realize it is the same in both cases. please refer to the \supp for detailed mathematical justification. We trained each GAN on CelebA and then used the pretrained model to fine tune the K-GAN according to KNN-FM on CelebA and each class from Zoo-faces as target Y. At each iteration, we search the KNN features of FC7 and FC6 in $\Psi_{\text{FC7}},\Psi_{\text{FC6}}$, which are the matrices of features of class Y and pick the mean of the two K features as two regularization targets for the GAN. Because of the use of the ball-tree search \cite{balltree} and SciPy library \cite{Scipy}, the search is done real-time and delay of the training time of K-GAN is only 70\% more compared to GAN with similar structure and for small target class Y of 100-500 instances. In general, the time complexity of the tree search is approximately $\mathcal{O}(s\log M)$, where $s$ is the feature dimensionality, and $M$ is the number of features in $\Psi$. The CycleGAN was trained independently on a subset of CelebA and each one of the 10 classes of Y and then fine tuned with samples of the first stage.This procedure was followed to obtain 10 K-DCGAN , 10 K-BEGAN , and 10 different CycleGANs corresponding to the 10 classes in Zoo-Faces. The coefficient of the KNN regularizer is picked to be 0.001 for FC7 and 0.0001 for FC6.



\subsection{Sampling}
After training, the models are cascaded, and sampling takes place in which we sample latent random vector $\mathbf{z} \in [-1,1]^{d}, d=100$ and pass it as input to the GANs and K-GANs similarly (No feature search happens at this stage). The image samples are then passed as input to the CycleGAN trained on the same X and Y sets that K-GAN was trained on. We obtain the output of an IAN model as in Fig \ref{qualititative} for K-BEGAN+CycleGAN. We show the closest training example next to some samples, as advised by \cite{evaluation}. More samples are provided in the  \supp.


\section{Results} \label{results-section}

\begin{table}[t]

\small
\tabcolsep=0.1cm

\centering
 \begin{tabular}{||c| c c c | c c c||} 
 \hline
 \multicolumn{1}{||c}{} & \multicolumn{3}{c}{\text{score}(.)} & \multicolumn{3}{c||}{$\text{err}(.)$}\\

 model  & X & Y & avg & X & Y & avg\\ [0.4ex] 
 \hline\hline
real(X) &100		& 0.02 &	50.01 &	0		&67.09	& 33.54 \\
real(Y) &10.38& 	63.29 &	36.84 &	55.33	&0		& 27.66 \\
real(X)+Cyc &12.24& 	44.03 &	28.13 &	48.57	&54.95	& 51.76 \\
 \hline
 DCGAN\cite{DCGAN}  &95.90& 	0.03 &	47.97 &	47.11	&64.23	& 55.67 \\
mxGAN &90.38& 	0.04 &	45.21 &	47.98	&64.40	& 56.19 \\
 BEGAN\cite{BEGAN}  &87.88 & 0.02 & 43.95 & 45.66 & 60.27 & 52.97  \\
P.GAN \cite{PerceptualSim} &\underline{97.69} & 	0.04 &	\underline{48.86} &	46.63	&63.10	& 54.87 \\
K-DCGAN &\textbf{98.77}& 	0.04 &	\textbf{49.40} &	46.97	&63.61	& 55.29 \\
K-BEGAN& 97.29 & 0.25 & 48.77 & \textbf{44.43} & 58.10 & 51.27 \\
\hline
rand+Cyc &1.44	& 0.08 &	0.76 &	49.42	&53.12	& 51.27 \\
DCGAN+Cyc &9.49	& 40.47 &	24.98 &	49.37	&53.51	& 51.44 \\
mxGAN+Cyc &9.75	& 39.11 &	24.43 &	49.97	&53.93	& 51.95 \\
P.GAN+Cyc&7.73	& 39.35 &	23.54 &	49.41	&53.42	& 51.41 \\
BEGAN+Cyc  & 30.05 & \underline{46.44} & 38.24 & 49.04 & \textbf{52.03} & \textbf{50.53}  \\
K-DCGAN+Cyc &10.19& 	39.62 &	24.90 &	49.12	& 52.89	& \underline{51.00} \\  
K-BEGAN+Cyc & 22.95 & \textbf{53.26} & 38.11 & 49.67 & \underline{52.80} & 51.23  \\ [1ex] 

 \hline 
 \end{tabular}
\caption{Objective Evaluation :The percentage scores (more is better) and normalized  pixel-error percentage (less is better) for base set X, and target set Y. \textbf{bold} is best and \underline{underlined} is second best. First three rows are actual data put for reference.The following six rows are different GANs, the last seven rows are cascaded GANs.Scores are averaged over the 10 different Y sets See Sec.\ref{evaluation} for details }
    \label{inception-table}
\end{table}

\subsection{Objective Evaluation} \label{evaluation}

\indent\textbf{Baselines:}\\
Ultimately, we want our generated samples to be indistinguishable from real samples; hence, we include the training images from X and Y as a reference for GAN model evaluation. We include the following GAN models (along with their cascade with CycleGAN) as baselines for fair comparison with our K-GAN and its cascade.

\noindent \textbf{DCGAN  ,  BEGAN:} Vanilla DCGAN \cite{DCGAN} and BEGAN\cite{BEGAN} trained on CelebA.

\noindent \textbf{mxGAN:} naive approach of combining X, and one class of Y by mixing the two sets as one training set of the DCGAN.

\noindent \textbf{PerceptualGAN:} feature matching GAN without the use of KNN search (random K selection), similar to \cite{PerceptualSim}. 

\noindent \textbf{rand+Cyc:} random noise translated with trained CycleGAN.

\begin{table}[t]
\tabcolsep=0.1cm

 \begin{tabular}{||c|| c|c|c|c|c|c|c||c||} 
 \hline
 model  & (1)   &(2)&(3) &(4) &(5) &(6)& \textbf{avg} & \text{HCA}\\ [0.5ex] 
 \hline
(1)	&\textbf{N/A} &	25.7	& 34.5	& 47.3	& 34.9	& 31.8	& 34.8 & \textbf{N/A}\\ \hline
(2)	&74.3 &\textbf{N/A}	& 38.8	& 38.8	& 36.90	& 4.0	& 47.2 & \textbf{N/A} \\ \hline
(3)	&65.5 &	61.20	& \textbf{N/A}	& 71.6	& 35.9	& 32.7	& 53.4 & 64.3 \\ \hline
(4)	&52.7 &	47.50	& 28.4	& \textbf{N/A}	& 24.8	& 20.7	& 34.8 & \textbf{68.3} \\ \hline
(5)	&65.1 &	63.10	& 64.1	& 75.2	& \textbf{N/A}	& 54.4	& \textbf{64}.4 & 58.0 \\ \hline
(6)	&68.2 &	53.00	& 67.3	& 79.3	& 45.60	&\textbf{N/A}	& 62.7 & 64.0 \\
    [1ex] 
 \hline 
 \end{tabular}
    \caption{Human Subjective Evaluation :Pair-wise quality comparison and Human Classification Accuracy (HCA) for six different models [(1):BEGAN,
 (2):K-BEGAN,
 (3):DCGAN+Cyc,
 (4):K-DCGAN+Cyc,
 (5):BEGAN+Cyc,
 (6):K-BEGAN+Cyc] . The Pair-wise percentages represent the frequency that a certain model's samples were preferred against their rival's model samples by human subjects ( the higher the more humans favoured that model) .HCA is for a classification task given for the subjects that were asked to classify the samples from different IANs ( guessing the target Y class from the imaginative sample ) .The higher the HCA the more easily identifiable are the samples. See Sec.\ref{subjective} for more details }
    \label{semi-results}
\end{table}
\textbf{Evaluation Metrics:} \\
As indicated by Theis \etal \cite{evaluation}, evaluating the generative models based on log likelihood's approximations (\eg Parzen window) should be avoided, and the models should be evaluated in the application they are being used in. To evaluate the imagination capability of our IAN model, we use the idea of inception score proposed by Salimans \etal \cite{inception-GANs2} on the target class Y. However, since we are developing different models for each class and the goal is not to model the data but to imagine new distributions between X and Y, we adopt the score for that purpose. We use the inception network\cite{inception-network} softmax probability of the modeled class directly (averaged over all samples), and then average the score over all models learned on the different classes and call this \textbf{score(Y)}. For base class assessment (the face), we use face detector accuracy as a metric. We used OpenCV \cite{opencv} implementation of the famous Viola-Jones detector \cite{viola-jones} to measure how our transformed faces retained the properties of the face. The face \textbf{score(X)} in table \ref{inception-table}  is defined as the percentage of images which the detector triggered a face. We use an extra evaluation criterion followed by \cite{PerceptualSim} and \cite{manifold-manipulation} by taking the normalized pixel-wise error defined as follows: $\text{err}(\mathbf{x}) =  \frac{\| \text{NN}(\mathbf{x})\|_{2}}{J}$
where $\text{NN}(\mathbf{x}) $ is the pixel-wise nearest neighbor to x in the training, $J$ is the average distance between uniform random noise in the training data. An $\text{err}(\mathbf{x})=1$ means our samples are no better than random noise, and a $\text{err}(\mathbf{x})=0$ indicates we generated a sample from training. For sampling, we follow a similar procedure to \cite{inception-GANs2} by using 1000 samples from different setups. We calculate the scores and error for both X and Y and the average of both on all the samples of each model for all ten classes of Y and the base class X. Table \ref{inception-table} summarizes the results.
\begin{figure}[t]
\includegraphics[width=1\linewidth]{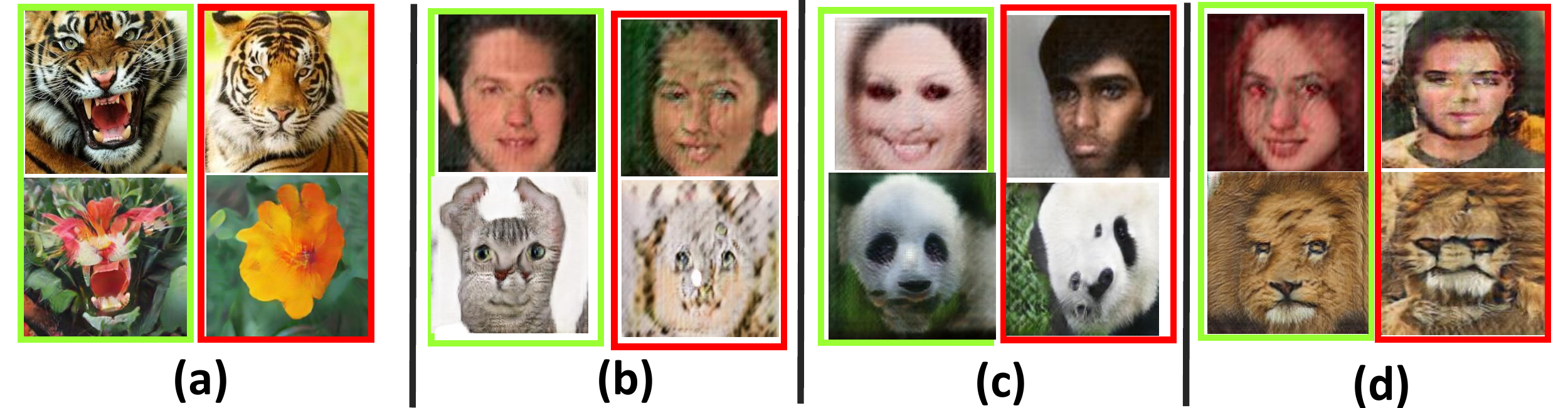}
\caption{\small \textbf{Ablation study}: \small \textbf{(a)} \textit{Hard Imagination}, trying to translate images from set X (\eg tiger) to very different set Y (\eg flower) using CycleGAN. \textbf{(b-d)}studying the effect of altering parts of the model on the GAN output and its corresponding cascaded sample with CycleGAN \textbf{(b)}abandoning the K-NN search produce high frequency artifacts.  \textbf{(c)}abandoning the KNN-FM amplifies blurriness after the cascade.\textbf{(d)}BEGAN vs DCGAN as base model. In all part, red is failure and green is success. }
\label{ablation2}
\end{figure}
\subsection{Subjective Evaluation} \label{subjective}
The nature of the task we are addressing, \ie imagination , is subjective and involves human judgment. Therefore , we did extensive human experiments using Amazon Mechanical Turk online tool. We designed two experiments to evaluate the  quality and the information conveyed by the samples from the different IAN models respectively.

\textbf{Quality Pair-wise Comparison}\\
In this experiment , the human subjects were shown human faces and a class from Zoo-Faces and then shown a pair of samples from two different setups . The task was to pick the one which was better looking and represent both the human face and the given class. A total of 15,000 pairs were shown to more than 150 unique human annotators to compare the quality of the generated samples from 6 different setups( 4 IANs and 2 GANs) with 15 combinations .The percentages shown in table \ref{semi-results} represent the frequency of human subjects picking that model against its rival , averaged over the 10 different classes of Zoo-faces. The higher the percentage of a model the better humans view the samples of that model.

\textbf{Human Subjective Classification Experiment}\\
In this experiment , the human subjects were asked to classify the samples generated from each IAN setup .Four IAN setups were trained on 10 different Y target classes from the Zoo-Faces dataset to give a total of 40 IAN models .A hundred samples ( like the ones in Fig. \ref{qualititative}) were drawn from each model to give a total of 4K images as a testing set for the annotators to classify. A total of 25 unique human participants classified the samples to the closest animal from Zoo-Faces classes. The Human Classification Accuracy (HCA) is shown in table \ref{semi-results} . The HCA indicates how easily identifiable are these generated samples. Please refer to \supp for the  Human Confusion Matrices of the four setups.
\subsection{Ablation Study }\label{ablation}

\textbf{Changing the Training Data}\\
We tried different data other than faces, and by using target class Y that are not animals and very far from the faces. We found that the more significant the difference is between X and Y the harder it becomes to produce meaningful results from the IAN structure. We show some cases of \textit{hard} imagination (\eg tiger to flower and shoe to tiger) in Fig \ref{ablation2}\textbf{(a)} with success (maintaining some key X properties after translation to Y) and failure (losing all X properties after translation). We reason that the further the classes in feature space, the possibility for the network to converge to a local minimum increases and hence not reaching to the target distribution.

\textbf{Abandoning the Search }\\
Picking a random image in the target set and using its features as the target of the regularizer seems to be a valid simpler option. However, randomly picking features can cause a rapid change in the objective of the GAN and hence introduce high-frequency artifacts as in Fig \ref{ablation2}\textbf{(b)}. These artifacts amplify when the samples pass by CycleGAN, and as results in table \ref{inception-table} suggest. Picking KNN features ensures stability in the objective and hence produce better samples.  

\textbf{Abandoning the Regularizer }\\
Using GAN to produce samples of the modeled data X then passing those samples to a trained cycleGAN to transform them and give the final samples seems intuitive. However, as we argued in section \ref{cascasde}, and as can be seen in Fig \ref{ablation2}\textbf{(c)}, using un-regularized GAN and building on top that can lead to systemic failure in which the error can amplify.

\textbf{Different GAN structures }\\
using the advanced BEGAN\cite{BEGAN} instead of DCGAN\cite{DCGAN} as base model produced better results from  resulting K-GAN and IANs. \ref{ablation2}\textbf{(d)}


\begin{figure}[t]
\begin{center}
  \includegraphics[width=1\linewidth]{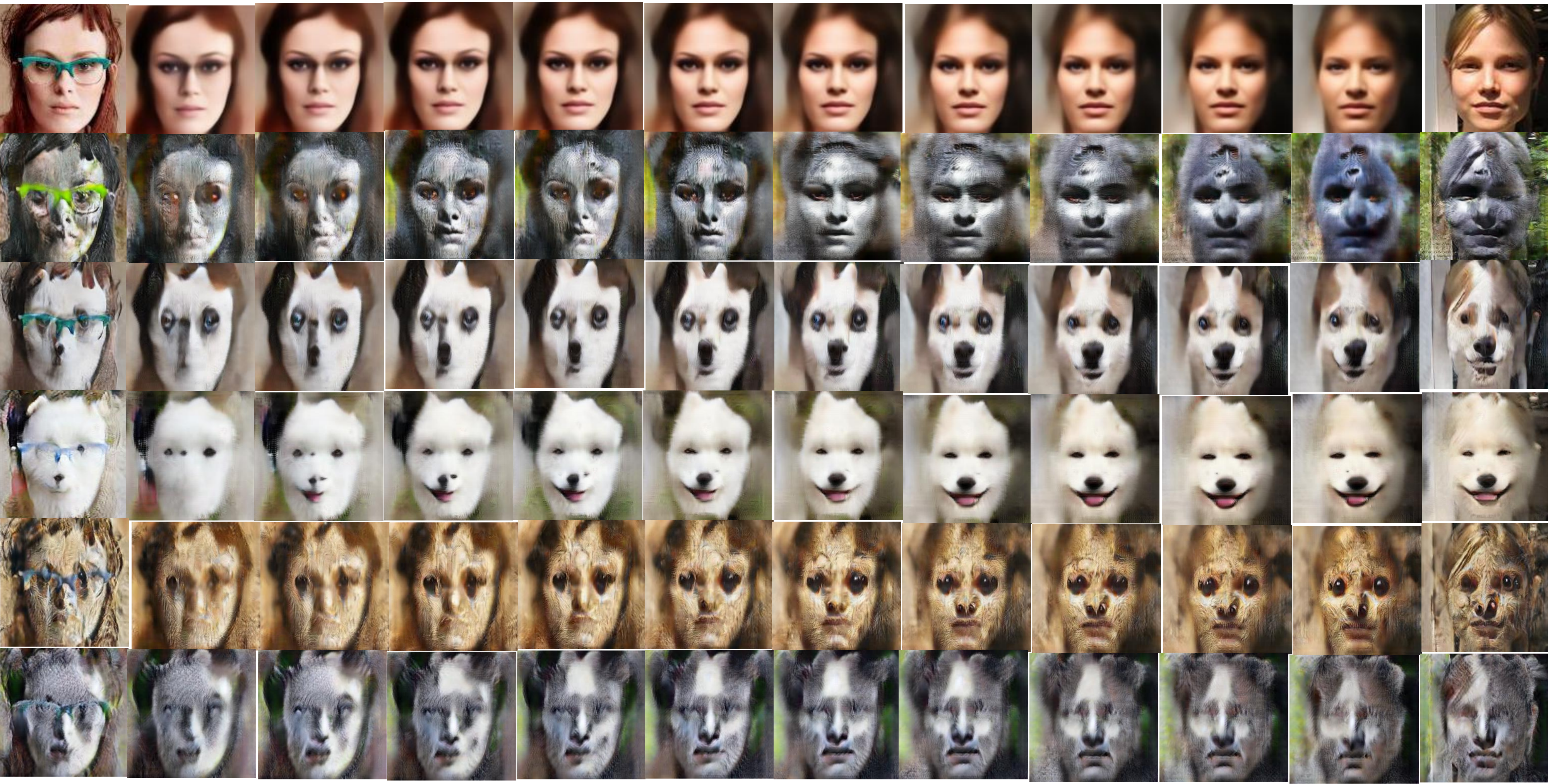}
  \end{center}
   \caption{\small \textbf{Multi-domain image manifold traversing}: moving from one point to another in different Y classes using the same pair in base set X . The top row is the actual traversing by convexing the codes obtained by the auto-encoder of BEGAN \cite{BEGAN} and then decoding the convex codes .The other rows are translations of the first row bu using different CycleGANs.}
\label{fig:traversing}
\end{figure}
\subsection{Discussion }\label{discussion}
Investigating the results in Fig \ref{qualititative} and tables \ref{inception-table} and \ref{semi-results} , we see that cascading any GAN model by CycleGAN increase the realism of target class Y by 40\% in the inception accuracy and be very close from the inception of an actual image. Also, it increases its likeability score in table \ref{semi-results} by about 15\%. However, it deteriorates the realism of the base class (the human face in our case). adding the K-NN regulizer made it easier for both Inception Network Softmax and humans HCA. The human likeability tends to favour unregularized IAN over regularized IAN , but K-GAN still outperforms GANs.


\section{Applications}
\subsection{Multi-Domain Image Manifold Traversing} \label{multi-traversing}
We follow the standard practice of utilizing the modeled distribution in manifold traversing \cite{manifold-manipulation}. But since we have trained different IANs on teh same X and different Y, we can do multi-traversing for different manifolds with only two images. First, we pick two training samples from X, \eg  $\mathbf{x}^{R}_{1},\mathbf{x}^{R}_{2}$. Then, we utilize the auto-encoder architecture of the discriminator of the trained BEGAN to encode the two samples and obtain their latent vectors$\mathbf{z}^{*}_{1},\mathbf{z}^{*}_{2}$  as follows: $ \mathbf{z}^{*} = \mathbf{D}_{\text{encoder}}(\mathbf{x}^{R}) $ 
Then, by taking $N$ convex points between these two latent vectors, we obtain $\mathbf{z}_{i}, i \in [1,\cdots,N]$ and finding the corresponding decoded  images as $\mathbf{D}_{\text{decoder}}(\mathbf{z}_{i})$. By translating these samples by CycleGAN, we can traverse different Y manifolds with only two samples from X as in Fig \ref{fig:traversing}. More qualitative results of this application are provided in the \supp.

\begin{figure}[t]
\begin{center}
\includegraphics[scale=0.3]{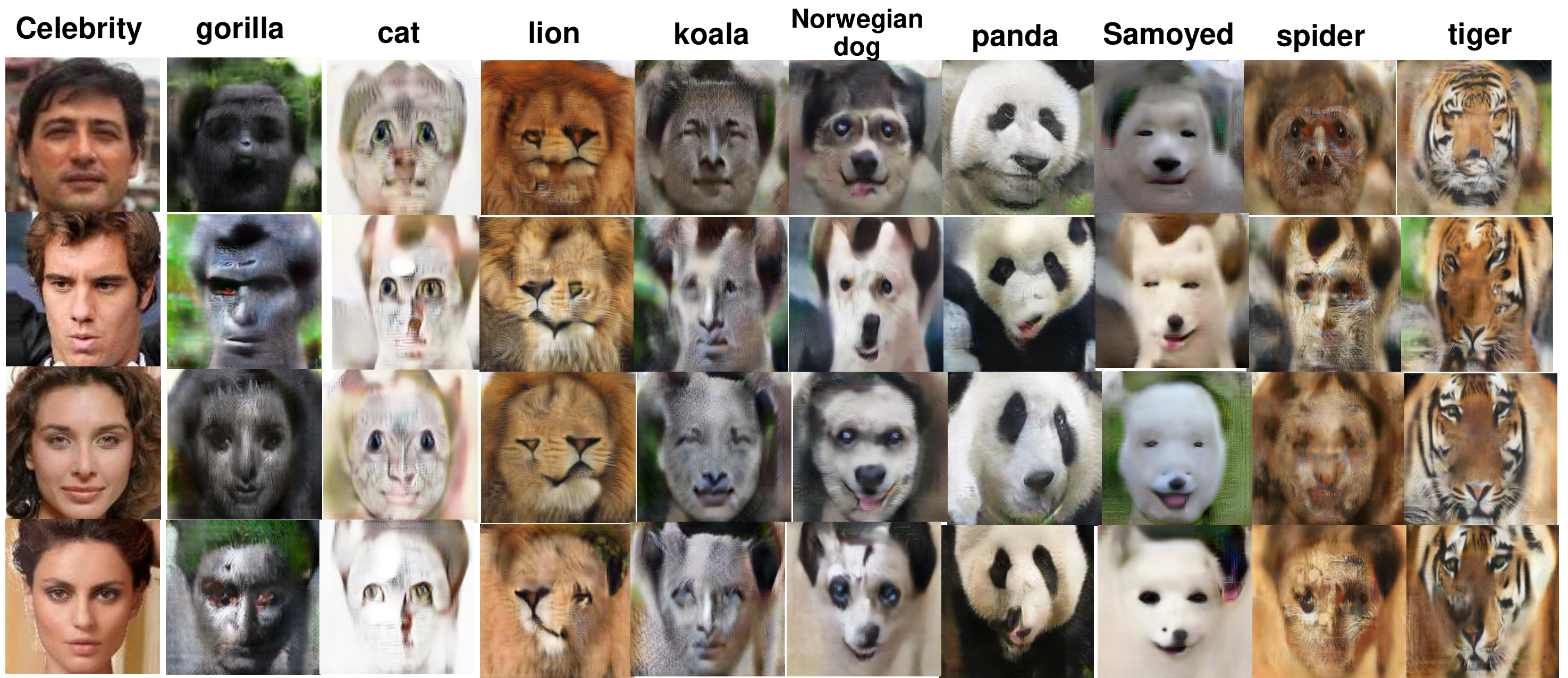}
\end{center}
   \caption{\small \textbf{Imaginative Faces:}: generating samples for humanoid animal characters from movie celebrities by using different IAN models trained in Sec.\ref{experemtn} can help in character design for movies and video games.}
\label{fig:zoo}
\end{figure}
\subsection{Generating Imaginative Faces to Help in Character Design}
Concept art plays an important role in character design in movies and video games. Automating creative concept art can be done by generating a lot of sketches for characters from which the artist would pick some and develop upon them the full model. This can reduce time and money in developing characters, and help the artist in generating creative concepts. To demonstrate this, we take a collection of celebrities from CelebA and encode them using the BEGAN Discriminator, as in Sec.\ref{multi-traversing}. Then, we pass the decoded instances to the CycleGAN for different animals from the Zoo-Faces dataset, obtaining animal counterparts for each celebrity in Fig \ref{fig:zoo}.  


\section{Conclusion and Future Work}
We have shown that using K-NN feature matching regularizer in GAN setup with two different sets (base X, target Y) helps in obtaining key properties of Y while keeping the GAN modeling of X with mathematical justification. We also, presented a framework for cascading GANs to address the task of imagining new distributions that combine base set X and target set Y .We  experiments with different IAN setups and show that while the regularized IAN comprising of K-GAN and CycleGAN was easily  identifiable by Inception networks and humans , humans favoured Vanilla IAN sometimes .We showed some applications like multi-domain manifold traversing.
Extending the IAN for more than two classes follow directly from our work. A potential use is to utilize IAN in zero-shot learning in which we learn from the generated samples based on some priors about the new class of unseen samples.




{\small
\bibliographystyle{ieee}
\bibliography{egbib}
}
\newpage \clearpage
\newpage
\appendix

\section{Deriving the KNN Loss}

We would like to show how to reach our final objective in (\ref{sup:main-final}) from the initial expression in (\ref{sup:main-initial})

\begin{equation}
\begin{aligned} 
 \min_{\mathbf{G}} \max_{\mathbf{D}}& ~ ~\mathbf{\mathit{L}}_{\text{GAN}}(\mathbf{G},\mathbf{D}_{X},\mathbf{P}_{X})  + \mu  ~ \mathbf{\mathit{L}}_{\text{KNN}}(\mathbf{G}, \mathbf{P}_{Y}),
\label{sup:main-final}
\end{aligned}
\end{equation}

\begin{equation}
\begin{aligned} 
 \min_{\mathbf{G}} \max_{\mathbf{D}}& ~ ~\mathbf{\mathit{L}}_{\text{GAN}}(\mathbf{G},\mathbf{D}_{X},\mathbf{P}_{X})  + \mu ~ \mathbb{E}_{\mathbf{z}\sim p_{\mathbf{z}}(\mathbf{z})} [H(\mathbf{G}(z) , \mathbf{P}_{Y})]
\label{sup:main-initial}
\end{aligned}
\end{equation}

\noindent for Generator $\mathbf{G}$ and discriminator $\mathbf{D}_{X}$ of modeled base set X with distribution $\mathbf{P}_{X}$ and latent vector $\mathbf{z} \in \mathbb{R}^{d}$,the GAN loss $\mathbf{\mathit{L}}_{GAN}(\mathbf{G},\mathbf{D}_{X},\mathbf{P}_{X})$ is given by 

\begin{equation} 
\begin{aligned}
 &
\mathbf{\mathit{L}}_{\text{GAN}}(\mathbf{G},\mathbf{D}_{X},\mathbf{P}_{X})=  \mathbb{E}_{\mathbf{x}\sim p_{x}(\mathbf{x})} [\log \mathbf{D}(x)] +  \mathbb{E}_{\mathbf{z}\sim p_{\mathbf{z}}(\mathbf{z})} [\log (1- \mathbf{D}(\mathbf{G}(z)))]
\end{aligned}
\label{sup:GAN}
\end{equation}
For Target set Y with distribution $\mathbf{P}_{Y} $, the K-nearest neighbor loss $\mathbf{\mathit{L}}_{\text{KNN}}(\mathbf{G},\mathbf{P}_{Y}) $ is defined as follows:
\begin{equation} \label{sup:K-GAN}
  \mathbf{\mathit{L}}_{\text{KNN}}(\mathbf{G},\mathbf{P}_{Y}) =  \mathbb{E}_{\mathbf{z}\sim p_{\mathbf{z}}(\mathbf{z})} \sum_{n=1}^{k} \| \mathbf{C}(\mathbf{G}(z)) - \mathbf{r}_{n}\left(\mathbf{C}\left(\mathbf{G}\left(z \right)\right) \right)  \|_{2}^{2}
\end{equation}
This is the distance between a generated sample from $\mathbf{G}$ by a latent vector $\mathbf{z}$ and its K nearest neighbors samples of the target distribution $\mathbf{P}_{Y}$. This distance in not pixels, but in high-level feature space defined by the Comparitor network $\mathbf{C}(\mathbf{b}) \in \mathbb{R}^{s}$ which takes an image $\mathbf{b} \in \mathbb{R}^{h \times w}$ as input and outputs feature vector . $\mathbf{r}_{n}(\mathbf{c})$ is the K-nearest neighbor function, a parametric order selection function that selects an element $\mathbf{y}_{i}$ from a set of $M-n+1$ elements based on how close it is to input vector 
$\mathbf{c}$. It can be described by the following:
\begin{equation}\label{sup:KNN}
    \mathbf{r}_{n}(\mathbf{c}) = \argmin_{\mathbf{y}_{i} \in \mathbb{\psi}_n} \| \mathbf{c} - \mathbf{y}_{i} \|_{2},
\end{equation}
where $\mathbb{\psi}_{n} $ is the set of deep features of $M-n+1$ images representing the target distribution $\mathbf{P}_{Y}$. The total number of samples features we have for $\mathbf{P}_{Y} $ is M , where the function  $\mathbf{r}_{n}(\mathbf{c})$ selects the nearest $n^{\text{th}}$ feature out of the remaining furthest  M-n+1   features after removing the nearest n-1 from the global set of all features $\mathbb{\Psi}$.

\noindent For the Cross-entropy in (\ref{sup:main-initial}) , we use the following definition of cross entropy H between two distributions $p,q$  as follows :
\begin{equation} \label{sup:xent}
  H(p ,q) = \mathbb{E}_{p} [-\log q ]
\end{equation}
To show the derivation , we first look into $\mathbf{P}_{Y} $. Since acquiring a full description of $\mathbf{P}_{Y}$ in deep feature space is infeasible in most cases, we settle to approximate it using $M$ target samples. We can get a proxy distribution of $\mathbf{P}_{Y}$ by using Kernel Density Estimation (KDE) on the deep features extracted by deep network $\mathbf{C}(\mathbf{x})$ from those $M$ samples in $\mathbf{P}_{Y}$. By picking Gaussian kernel for the KDE, the proxy distribution estimator $\hat{\mathbf{P}}_{Y,\text{proxy}}$ is defined by a Bayesian non-parametric way at any point $\mathbf{C}(\mathbf{b}) \in \mathbb{R}^{s}$ for $\mathbf{b \in \mathbb{R}^{h \times w}}$ as follows:

\begin{equation} \label{sup:proxy}
\hat{\mathbf{P}}_{Y,\text{proxy}}(\mathbf{b}) = \frac{1}{M\sigma \sqrt{2\pi}}\sum_{i=1}^{M} \exp\left(-\frac{\|\mathbf{C}(\mathbf{b}) - \mathbf{C}(\mathbf{y}_{i})\|_2^2}{\sigma^2}\right)  
\end{equation}
\noindent $\mathbf{y}_{i}$ is the $i^{\text{th}}$ sample describing $\mathbf{P}_{Y}$ and $\sigma^{2}$ is the Gaussian kernel variance. 
For far images from the $M$ samples of  $\hat{\mathbf{P}}_{Y,proxy}$ the probability becomes exponentially negligible. Hence, we we can investigate the effect of a subset of K points (out of the M points $\mathbf{C} (\mathbf{y}_{i})$) that are closest to $\mathbf{C}(\mathbf{b})$ to compute (\ref{sup:proxy}). We replace $M$ by $K$ in (\ref{sup:proxy}), and pick $\sigma = 1$ for simplicity to get the following  :

\begin{equation} \label{sup:proxy2}
\hat{\mathbf{P}}_{Y,\text{proxy}}(\mathbf{b}) \approx \frac{1}{M \sqrt{2\pi}}\sum_{i=1}^{K} \exp\left(-\|\mathbf{C}(\mathbf{b}) - \mathbf{r}_{i}(\mathbf{C}(\mathbf{b}))\|_2^2\right)  
\end{equation}
\noindent where $\mathbf{r}_{i}$ is described in (\ref{sup:KNN}). We use a finite discrete special form of the Jansen inequality described by Theorem 7.3 of (\cite{Jansen}) as follows\\

\begin{equation} \label{sup:Jensen}
f(\sum_{i=1}^{n}u_{i}\lambda_{i}) ~~\leq ~~  \sum_{i=1}^{n}f(u_{i})\lambda_{i}
\end{equation}
\noindent where $f $ is convex function for any $u_{i}$ , $\lambda_{i}$ are set of $n$ weights with $\sum_{i=1}^{n}\lambda_{i} = 1$ .\\By Picking $ f(\mathbf{x}) =  \exp(-\mathbf{x})$ a convex function defined in$ [0,\infty)^{d}$, and picking $u_{i}(\mathbf{b}) = \| \mathbf{C}(\mathbf{b}) - \mathbf{r}_{i}(\mathbf{C}(\mathbf{b})) \|_{2}^{2}$ for $\mathbf{b} \in \mathbb{R}^{h\times w}, \mathbf{r}_{i}$ is just like in (\ref{sup:KNN}) and $\mathbf{C}$ is like in (\ref{sup:K-GAN}), and picking $\lambda_{i} = \frac{1}{K}$ ,and  $n=K$, inequality (\ref{sup:Jensen}) becomes :

\begin{equation} 
\begin{aligned}
\exp(&-\frac{1}{K}\sum_{i=1}^{K} \| \mathbf{C}(\mathbf{b}) - \mathbf{r}_{i}(\mathbf{C}(\mathbf{b})) \|^{2}_{2}) \\ &\leq~ \frac{1}{K}\sum_{i=1}^{K} \exp( -\| \mathbf{C}(\mathbf{b}) - \mathbf{r}_{i}(\mathbf{C}(\mathbf{b})) \|^{2}_{2} ) 
\end{aligned}
\label{sup:Jensensubs}
\end{equation}
Noting that the right-hand side of (\ref{sup:Jensensubs}) is scaled version of the proxy distribution approximate  $\hat{\mathbf{P}}_{Y,\text{proxy}}(\mathbf{b}) $ described in (\ref{sup:proxy2}), we reach to the following inequality

\begin{equation} 
\begin{aligned}
\exp(&-\frac{1}{K}\sum_{i=1}^{K} \| \mathbf{C}(\mathbf{b}) - \mathbf{r}_{i}(\mathbf{C}(\mathbf{y}_{i})) \|^{2}_{2}) \\  &\leq~ \frac{M\sqrt{2\pi}}{K}\hat{\mathbf{P}}_{Y,\text{proxy}}(\mathbf{b})
\end{aligned}
\label{sup:Jensensubs2}
\end{equation}

\noindent Taking the natural logarithm ( a monotonically non-decreasing function)  of both sides of (\ref{sup:Jensensubs2}) and then negating both sides results in the following inequality :\\  

\begin{equation} 
\begin{aligned}
-&\log \left (\frac{M\sqrt{2\pi}}{K}\hat{\mathbf{P}}_{Y,\text{proxy}}(\mathbf{b})\right)  ~ \leq~ \frac{1}{K}\sum_{i=1}^{K} \| \mathbf{C}(\mathbf{b}) - \mathbf{r}_{i}(\mathbf{C}(\mathbf{y}_{i}) \|^{2}_{2}) 
\end{aligned}
\label{sup:bound}
\end{equation}
\noindent Rearranging (\ref{sup:bound}) to get the following:

\begin{equation} 
\begin{aligned}
-&\log \hat{\mathbf{P}}_{Y,\text{proxy}}(\mathbf{b}) ~\leq~  \log \frac{M\sqrt{2\pi}}{K}  +  \frac{1}{K}\sum_{i=1}^{K} \| \mathbf{C}(\mathbf{b}) - \mathbf{r}_{i}(\mathbf{C}(\mathbf{y}_{i}) \|^{2}_{2}) 
\end{aligned}
\label{sup:bound2}
\end{equation}

\noindent By replacing the $\mathbf{b}$ by the output of the generator $ \mathbf{G}(\mathbf{z})$ sampled from latent vector $\mathbf{z} $ and taking the expectation, we get the following expected upper bound

\begin{equation} 
\begin{aligned}
\mathbb{E}&_{\mathbf{z}\sim p_{\mathbf{z}}(\mathbf{z})}[-\log \hat{\mathbf{P}}_{Y,\text{proxy}}(\mathbf{\mathbf{G}(\mathbf{z})})]  \\ \leq & \log \frac{M\sqrt{2\pi}}{K}   ~+~\mathbb{E}_{\mathbf{z}\sim p_{\mathbf{z}}(\mathbf{z})}[ \frac{1}{K}\sum_{i=1}^{K} \| \mathbf{C}(\mathbf{G}(\mathbf{z})) - \mathbf{r}_{i}(\mathbf{C}(\mathbf{y}_{i}) \|^{2}_{2})] 
\end{aligned}
\label{sup:bound3}
\end{equation}
\noindent We note that the left side is the cross entropy between $\mathbf{G}$,$\mathbf{P}_{Y}$, and the second term in the right hand side is scaled version of the KNN loss  $\mathbf{\mathit{L}}_{\text{KNN}}(\mathbf{G},\mathbf{P}_{Y})$ in (\ref{sup:K-GAN}),to get :

\begin{equation} 
\begin{aligned}
~\mathbb{E}_{\mathbf{z}\sim p_{\mathbf{z}}(\mathbf{z})}& [H(\mathbf{G}(z), \mathbf{P}_{Y})]  ~~ \leq ~~ \log \frac{M\sqrt{2\pi}}{K}  + \frac{M\sqrt{2\pi}}{K}\mathbf{\mathit{L}}_{\text{KNN}}(\mathbf{G},\mathbf{P}_{Y})
\end{aligned}
\label{sup:bound-entropy}
\end{equation}

\noindent The expression in (\ref{sup:bound-entropy}) gives upper bound for the cross entropy using the KNN loss. By minimizing the KNN loss for parameters of $\mathbf{G}$ as in (\ref{sup:main-final}) we insure minimizing the cross entropy $\mathbb{E}_{\mathbf{z}\sim p_{\mathbf{z}}(\mathbf{z})} [H(\mathbf{G}(z), \mathbf{P}_{Y})]$ as in (\ref{sup:main-initial}) and hence establish mathematical justification for using KNN loss instead of cross entropy.

\clearpage

\section{Justifying regressing to the mean of the K-features instead of each independently during the training of K-GAN}
We would like to show that the equivalence of :
\begin{equation} \label{eq:regression-all}
  \argmin_{\mathbf{w}}~~ \mathbb{E}_{\mathbf{z}\sim p_{\mathbf{z}}(\mathbf{z})} \sum_{i=1}^{k} \| \mathbf{C}(\mathbf{G}(\mathbf{z},\mathbf{w})) - \mathbf{d}_{i}  \|_{2}^{2}
\end{equation}
\begin{equation} \label{eq:regresswion-mean}
  \argmin_{\mathbf{w}}~~ \mathbb{E}_{\mathbf{z}\sim p_{\mathbf{z}}(\mathbf{z})}  \left\| \mathbf{C}(\mathbf{G}(\mathbf{z},\mathbf{w})) - \frac{1}{k}\sum_{i=1}^{k}\mathbf{d}_{i}  \right\|_{2}^{2}
\end{equation}
  \noindent where $\mathbf{C}$ is the feature extraction network ,$\mathbf{G}(\mathbf{z},\mathbf{w})$ is the Generator network with parameters $\mathbf{w}$ and random sampling latent vector $\mathbf{z}$ sampled from uniform distribution, and $\mathbf{d}_{i}$'s are the $K$ chosen features in the time step t. \\
  By starting from Eq.(\ref{eq:regression-all}) and using the notation $\mathbf{f}(\mathbf{w})$ instead of $\mathbf{C}(\mathbf{G}(\mathbf{z},\mathbf{w}))$,and $\mathbf{\bar{d}}$ as the mean $\frac{1}{k} \sum_{i=1}^{k}\mathbf{d}_{i}$, Eq.(\ref{eq:regression-all}) can be expanded as the following 
  \begin{equation} 
\begin{aligned}
  &\argmin_{\mathbf{w}}~~  \sum_{i=1}^{k}\| \mathbf{f}(\mathbf{w}) - \mathbf{d}_{i}  \|_{2}^{2} \\
  =  &\argmin_{\mathbf{w}}~~ \frac{1}{k} \sum_{i=1}^{k}\| \mathbf{f}(\mathbf{w}) - \mathbf{d}_{i}  \|_{2}^{2} \\
   = &\argmin_{\mathbf{w}}~~ \frac{1}{k} \sum_{i=1}^{k}\left<\mathbf{f}(\mathbf{w}),\mathbf{f}(\mathbf{w})\right> -  \frac{2}{k}\sum_{i=1}^{k} \left<\mathbf{f}(\mathbf{w}),\mathbf{d}_{i}\right> + \frac{1}{k} \sum_{i=1}^{k}\left<\mathbf{d}_{i},\mathbf{d}_{i}\right> \\ 
   = &\argmin_{\mathbf{w}}~~  \left<\mathbf{f}(\mathbf{w}),\mathbf{f}(\mathbf{w})\right> - 2\left<\mathbf{f}(\mathbf{w}),\mathbf{\bar{d}}\right> + \frac{1}{k} \sum_{i=1}^{k}\left<\mathbf{d}_{i},\mathbf{d}_{i}\right> \\
      = &\argmin_{\mathbf{w}}~~  \left<\mathbf{f}(\mathbf{w}),\mathbf{f}(\mathbf{w})\right> - 2\left<\mathbf{f}(\mathbf{w}),\mathbf{\bar{d}}\right> 
\end{aligned}
\label{eq:regression-expanded}
\end{equation}
where $\left<.,.\right> $ is the inner product and after utilizing the property $\sum_{i}\left<\mathbf{a},\mathbf{d}_{i}\right> = \left<\mathbf{a}, \sum_{i} \mathbf{d}_{i}\right> $. \\
  On the other hand, Eq.(\ref{eq:regresswion-mean}) can be expanded ( with the new notation )  as the following
  
    \begin{equation} 
\begin{aligned}
  &\argmin_{\mathbf{w}}~~   \| \mathbf{f}(\mathbf{w}) - \mathbf{\bar{d}}  \|_{2}^{2} \\
  =  &\argmin_{\mathbf{w}}~~ \left<\mathbf{f}(\mathbf{w}),\mathbf{f}(\mathbf{w})\right> - 2\left<\mathbf{f}(\mathbf{w}),\mathbf{\bar{d}}\right> +\left<\mathbf{\bar{d}},\mathbf{\bar{d}}\right> \\
   = &\argmin_{\mathbf{w}}~~ \left<\mathbf{f}(\mathbf{w}),\mathbf{f}(\mathbf{w})\right> - 2\left<\mathbf{f}(\mathbf{w}),\mathbf{\bar{d}}\right>  
\end{aligned}
\label{eq:regression-expanded2}
\end{equation}
\indent We can see that the final expressions in Eq.(\ref{eq:regression-expanded}) and Eq.(\ref{eq:regression-expanded2}) are equivalent. Hence , the equivalence between  Eq.(\ref{eq:regression-all}) and Eq.(\ref{eq:regresswion-mean}) is established , and picking the mean of the K-NN features at each iteration is mathematically justified , while improving the speed of the training of the K-GAN.
  \clearpage

\section{Imaginative Face Generation for Character Design Aid }
We show more examples of celebrity zoo in the following page.

\begin{figure*}
\begin{center}
\includegraphics[width = \linewidth]{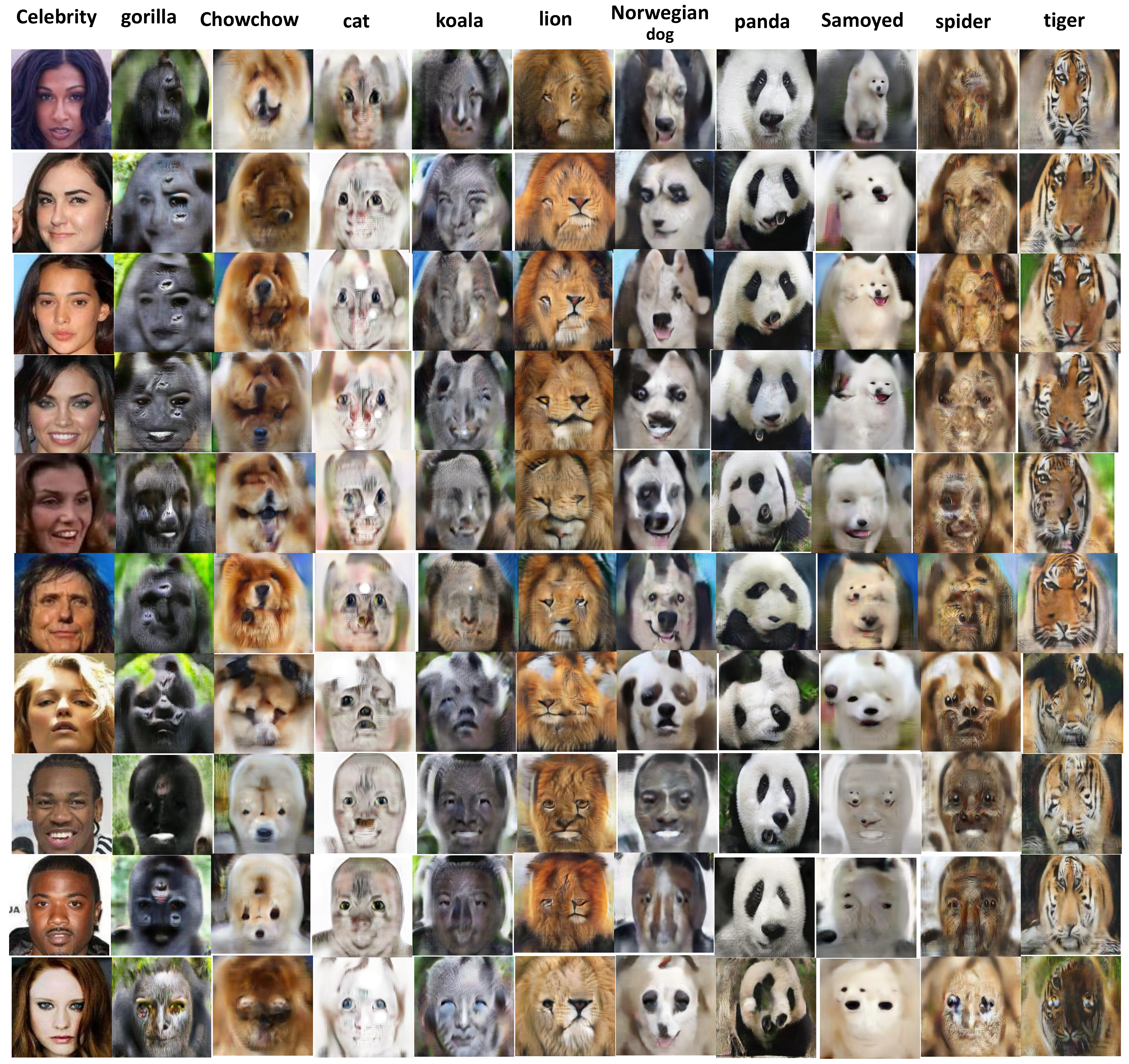}
\end{center}
   \caption{ \textbf{Imaginative Faces}: using different IAN models trained in Sec.4}
\label{sup:zoo1}
\end{figure*}

\section{image Manifold Traversing}
We do image manifold traversing for more classes.
\begin{figure*}
\begin{center}
\includegraphics[width = \linewidth]{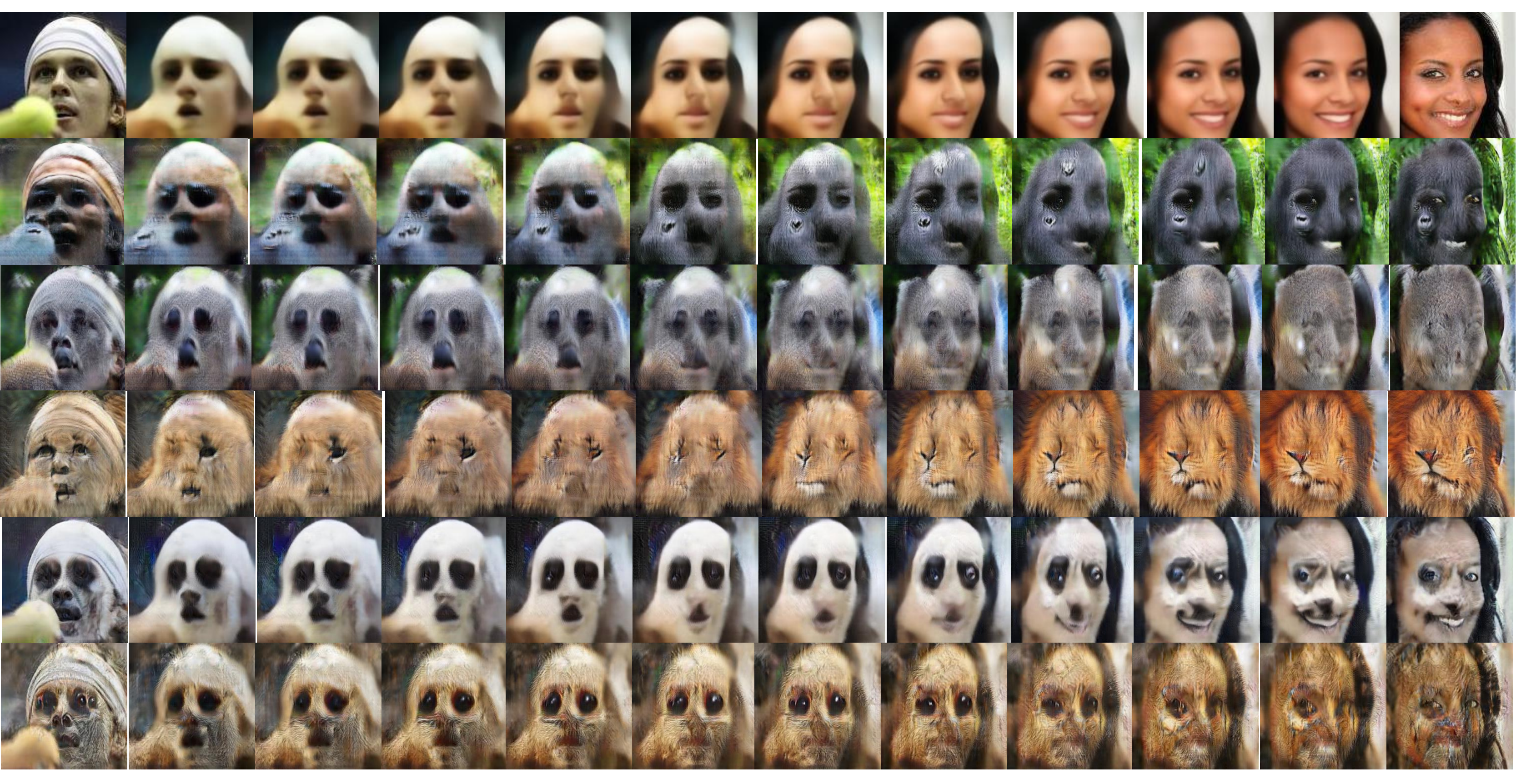}
\end{center}
   \caption{ \textbf{multi-domain image manifold traversing}: Multi-domain image manifold traversing: moving from one point to another in different Y classes using the same pair in base set X}
\label{sup:trav-1}
\end{figure*}

\begin{figure*}
\begin{center}
\includegraphics[width = \linewidth]{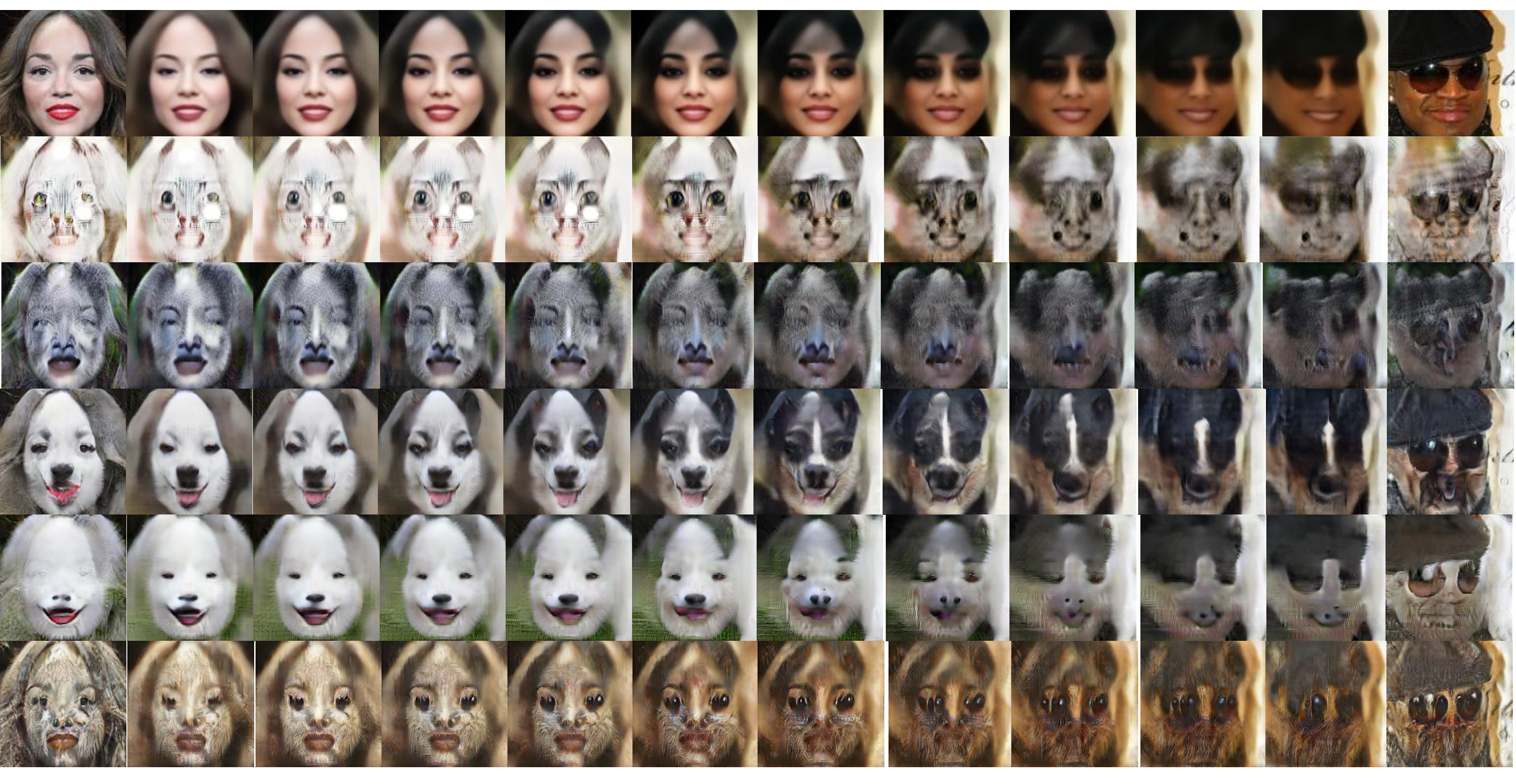}
\end{center}
   \caption{ \textbf{multi-domain image manifold traversing}: Multi-domain image manifold traversing: moving from one point to another in different Y classes using the same pair in base set X}
\label{sup:trav-2}
\end{figure*}

\begin{figure*}
\begin{center}
\includegraphics[width = \linewidth]{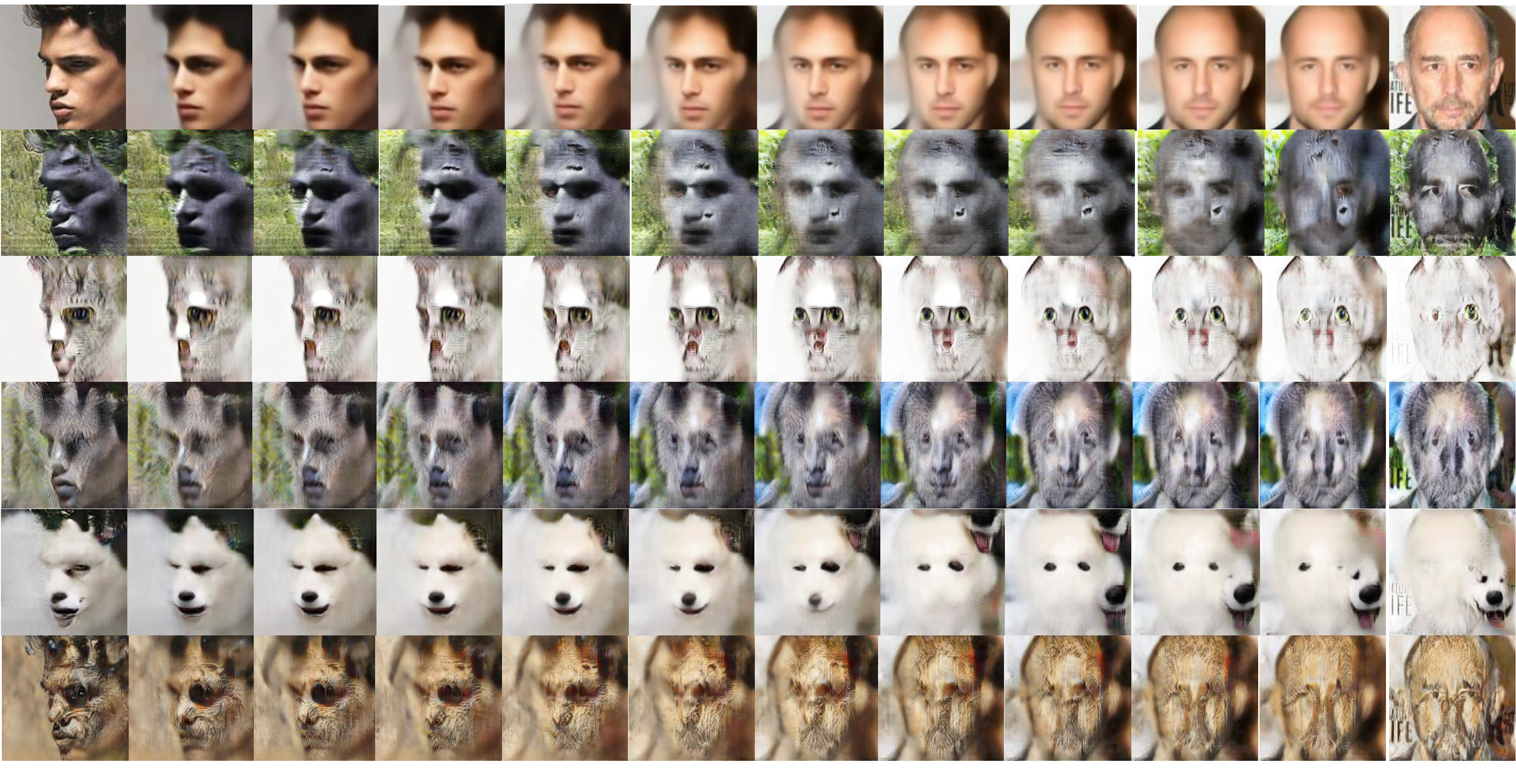}
\end{center}
   \caption{ \textbf{multi-domain image manifold traversing}: Multi-domain image manifold traversing: moving from one point to another in different Y classes using the same pair in base set X}
\label{sup:trav-3}
\end{figure*}

\begin{figure*}
\begin{center}
\includegraphics[width = \linewidth]{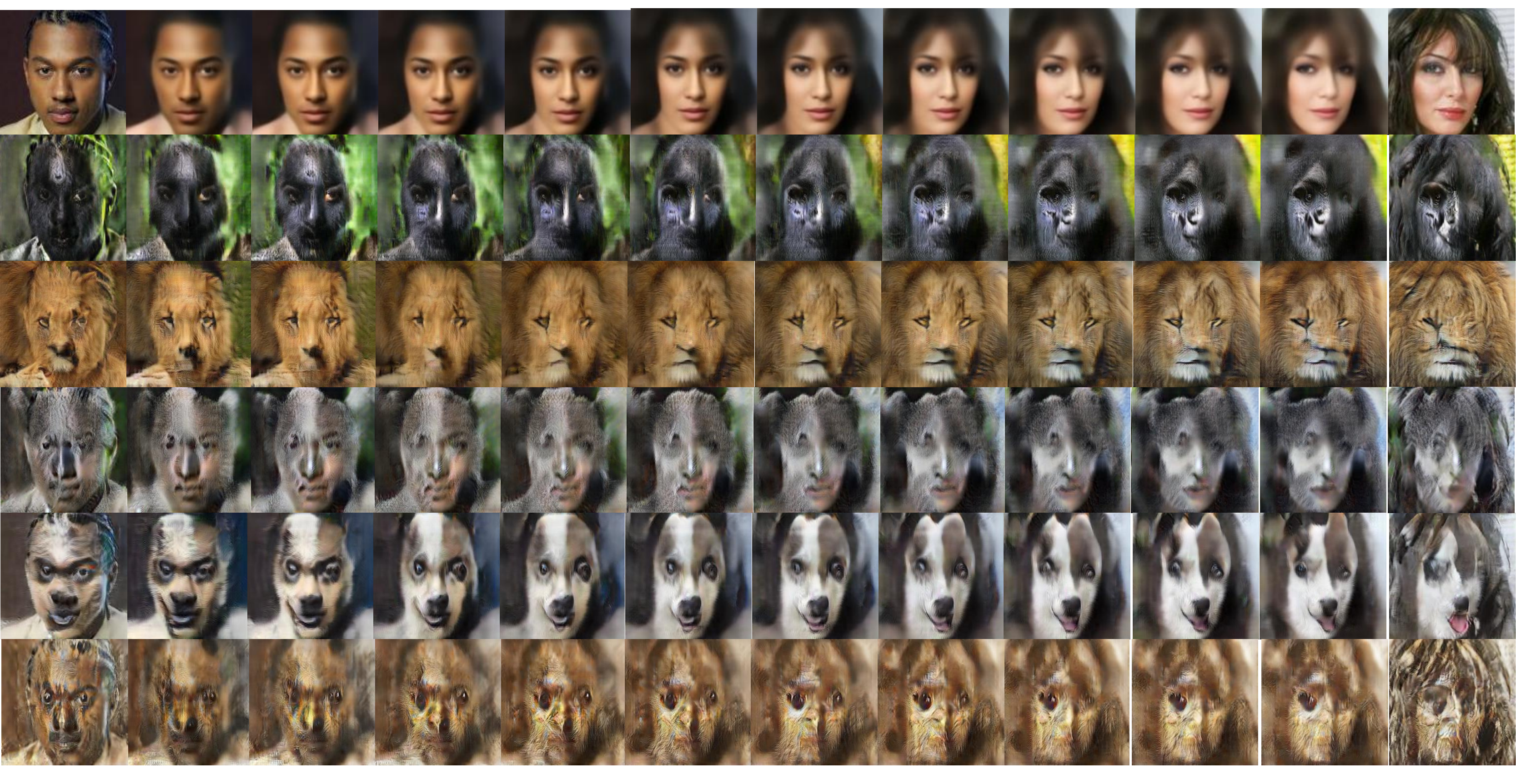}
\end{center}
   \caption{ \textbf{multi-domain image manifold traversing}: Multi-domain image manifold traversing: moving from one point to another in different Y classes using the same pair in base set X}
\label{sup:trav-4}
\end{figure*}

\begin{figure*}
\begin{center}
\includegraphics[width = \linewidth ]{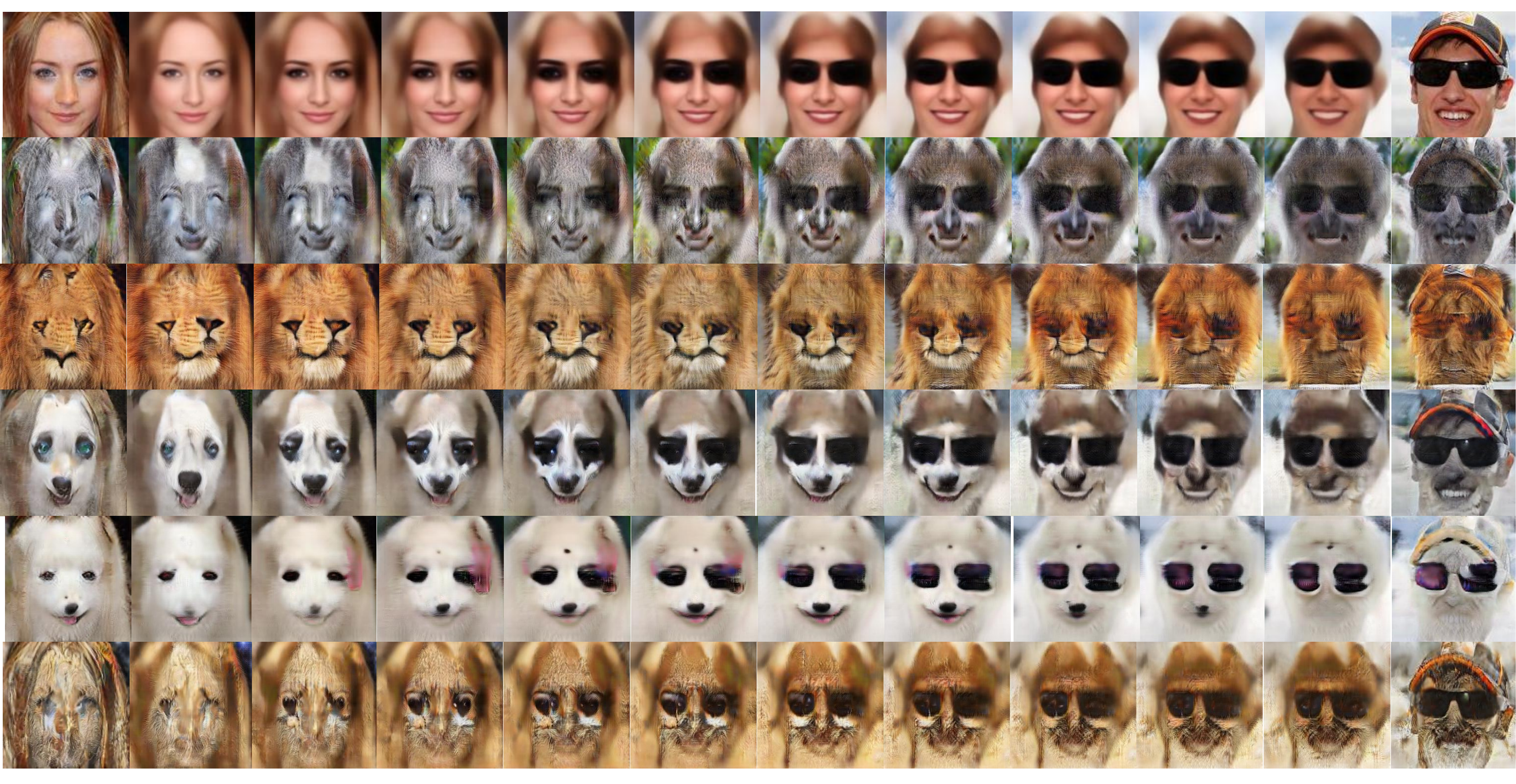}
\end{center}
   \caption{ \textbf{multi-domain image manifold traversing}: Multi-domain image manifold traversing: moving from one point to another in different Y classes using the same pair in base set X}
\label{sup:trav-5}
\end{figure*}

\clearpage

\section{Selecting the Parameter K and $\mu$ for Training K-GAN} 
To study the effect of changing the K parameter of nearest neighbor search, we look at the cardinality of the set $\chi$ (a subset of the set $\Psi$ of a high level feature of the target $Y$). $\chi$ contains different features from class $Y $ that have been selected during the KNN search in K-GAN training. A bigger $\chi$ cardinality is an indication of a better hyper-parameter selection, which means we have more effectively utilized the set $\Psi$ and the KNN search did not pick one feature several times during the training. We use this number as a validation metric to assess the quality of the training. We picked $K=4$ because it compromises both the utilization of $\Psi$ and the stability of the training. The value of $\mu$ hyperparameter (coefficient of the KNN regularizer) was picked to be 0.001 for FC7 and 0.0001 for FC6. For $\chi$ the subset of features picked by KNN search during the training of K-GAN out of M features, the following Fig \ref{K-selection} shows percentage of card($\chi$) to $M$ for different choice of target class $Y$ and different values of hyper-parameter $K$. We used this percentage as quick validation metric to assess the quality of the training. We picked $K $ such that we covers large percentage of the target set while preserving the uniqueness of each image feature. Also , since we are taking the mean of the KNN features , larger K is less prone to noise especially if there is an image in the training that is very unique from others and can be picked by the KNN search several times.\\
The $\mu_{\text{FC7}},\mu_{\text{FC7}}$ (the coefficients of FC7 and FC6 regularizers of K-GAN  ) were picked to insure that the key properties were indeed transferred from Y to X, at the same time not over-regularizing the model which can lead to the collapse of the K-GAN model and producing only a single image. Table \ref{parameters} shows the values of $K$ and $\mu_{\text{FC7}},\mu_{\text{FC7}}$ picked to train the final 10 K-GAN models that we reported its samples and did all the tests on for the 10 different Zoo-Faces dataset classes. 

  \begin{figure}[h!]
\begin{center}
  \includegraphics[width=0.6\linewidth]{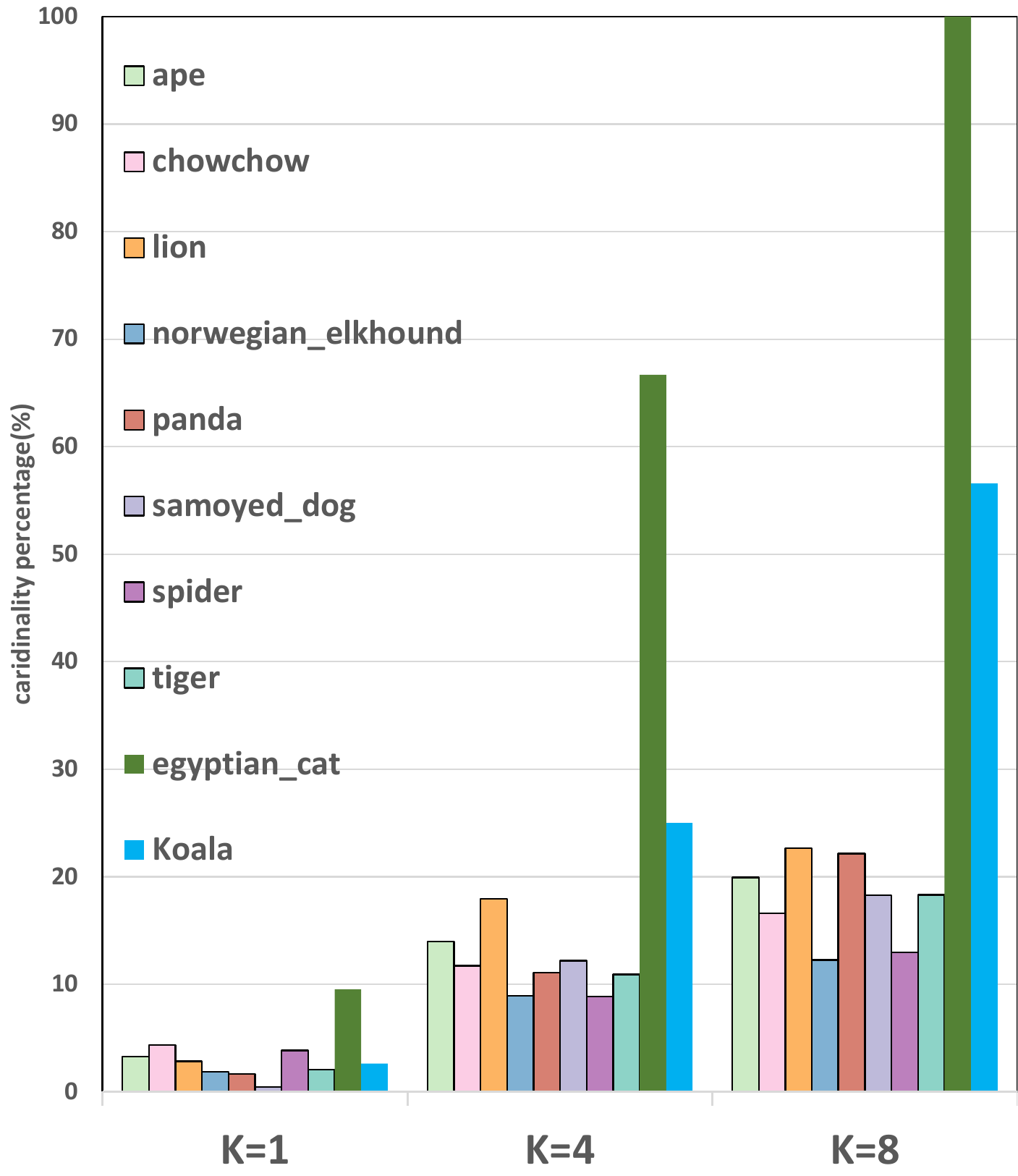}
\end{center}
  \caption{ The percentage of used features in training K-GAN model with K= 1,4,and 8 , and using different target classes $Y $   }
\label{K-selection}
\end{figure}

 \begin{table}[]
\centering
 \begin{tabular}{||c| c c c||} 
 \hline
 K-GAN model  & K & $\mu_{\text{FC7}}$ & $\mu_{\text{FC6}}$ \\ [0.5ex] 
 \hline\hline
  Apes  & 8 & 0.012 &0.001 \\ 
  egyptian-cat  & 8 & 0.010 &0.001 \\ 
Chowchow & 8 &  0.012 &0.001\\
  koala & 4 & 0.010 & 0.001   \\
  Lion  & 8 & 0.010 &0.001 \\ 
Norwegian elkhound  & 4 & 0.012 &0.001 \\ 
  panda  & 4 & 0.010 &0.001 \\ 
   samoyed dog & 4 & 0.010 & 0.001   \\
    spider & 8 & 0.010 & 0.001   \\
  tiger  & 4 & 0.010 &0.001 \\ [1ex] 
 \hline 
 \end{tabular}
    \caption{The values of parameter K and coefficients of regularizer used in training different K-GAN models. }
    \label{parameters}
\end{table}
 

\clearpage

\section{Samples of the Zoo-Faces Dataset}
  \begin{figure}[h]
\begin{center}
  \includegraphics[width=0.5\linewidth]{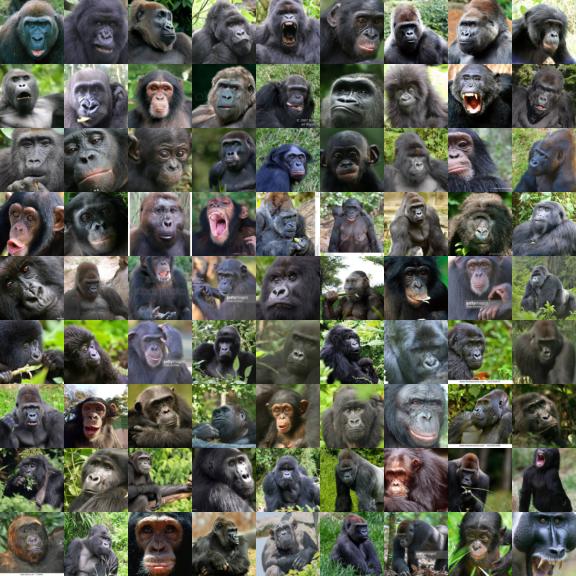}
\end{center}
  \caption{ Samples of the Zoo-Faces dataset  }
\label{fig:dataset-1}
\end{figure}

  \begin{figure}[h]
\begin{center}
  \includegraphics[width=0.5\linewidth]{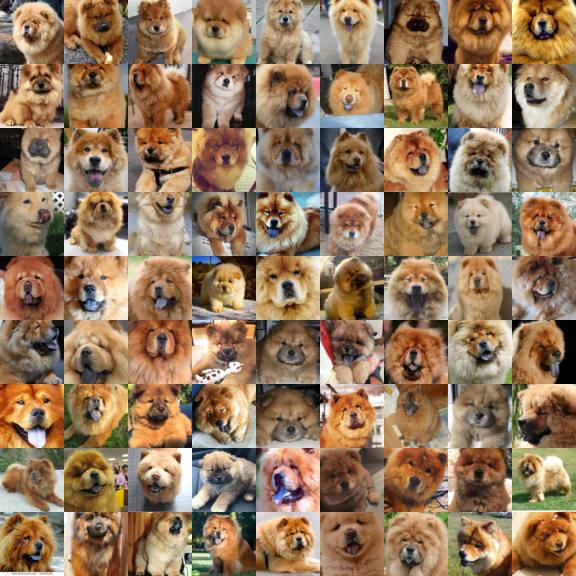}
\end{center}
  \caption{ Samples of the Zoo-Faces dataset  }
\label{fig:dataset-2}
\end{figure}

  \begin{figure}[h]
\begin{center}
  \includegraphics[width=0.5\linewidth]{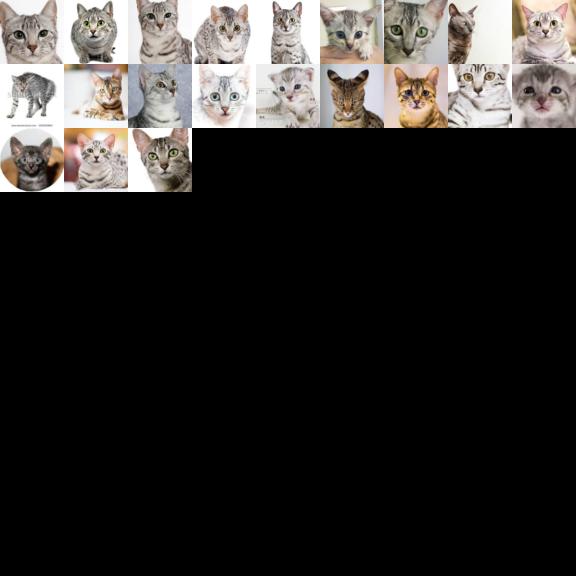}
\end{center}
  \caption{ Samples of the Zoo-Faces dataset   }
\label{fig:dataset-3}
\end{figure}

  \begin{figure}[h]
\begin{center}
  \includegraphics[width=0.5\linewidth]{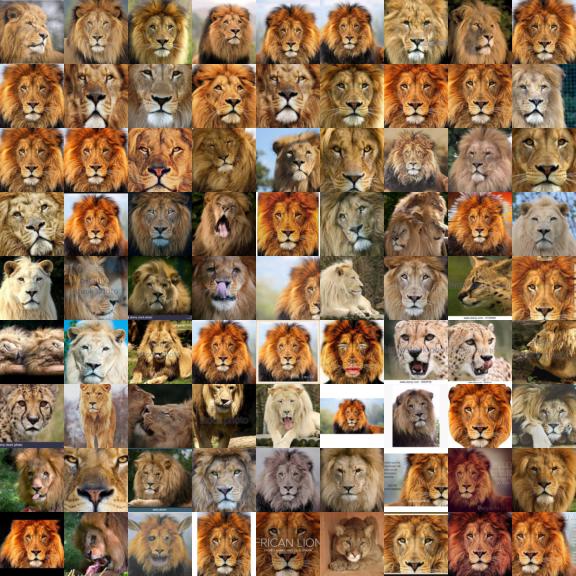}
\end{center}
  \caption{ Samples of the Zoo-Faces dataset   }
\label{fig:dataset-4}
\end{figure}

  \begin{figure}[h]
\begin{center}
  \includegraphics[width=0.5\linewidth]{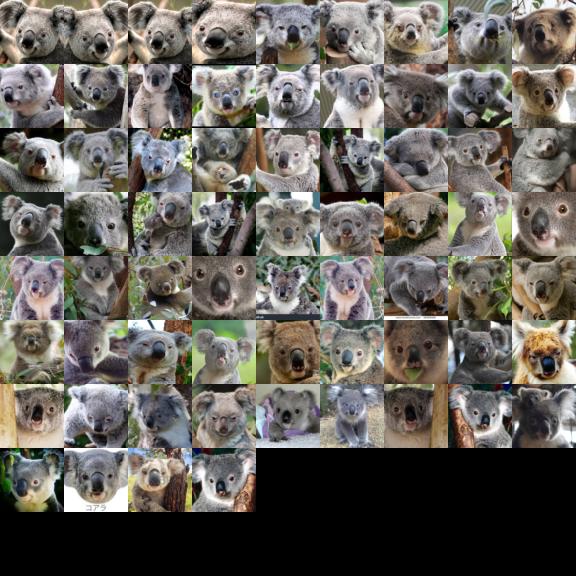}
\end{center}
  \caption{Samples of the Zoo-Faces dataset   }
\label{fig:dataset-5}
\end{figure}

  \begin{figure}[h]
\begin{center}
  \includegraphics[width=0.5\linewidth]{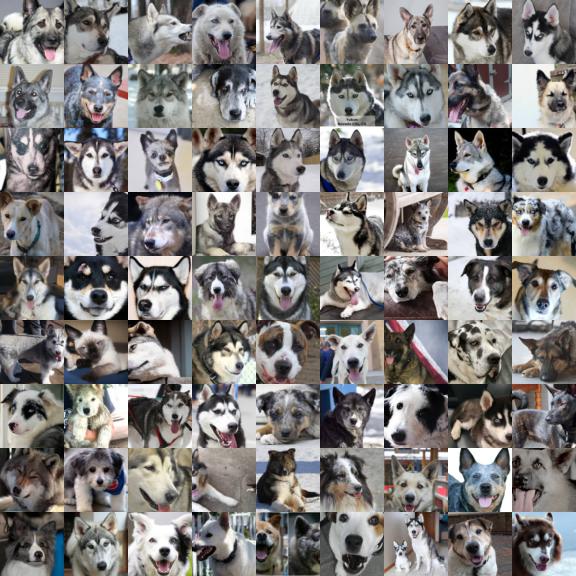}
\end{center}
  \caption{ Samples of the Zoo-Faces dataset  }
\label{fig:dataset-6}
\end{figure}
  
  \begin{figure}[h]
\begin{center}
  \includegraphics[width=0.5\linewidth]{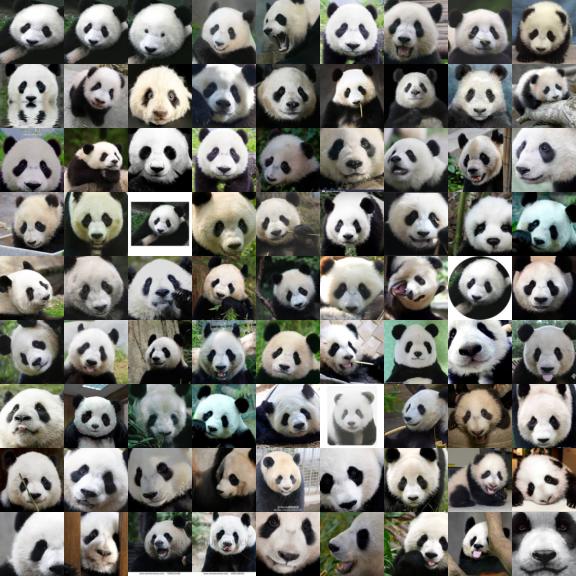}
\end{center}
  \caption{ Samples of the Zoo-Faces dataset   }
\label{fig:dataset-7}
\end{figure}

  \begin{figure}[h]
\begin{center}
  \includegraphics[width=0.5\linewidth]{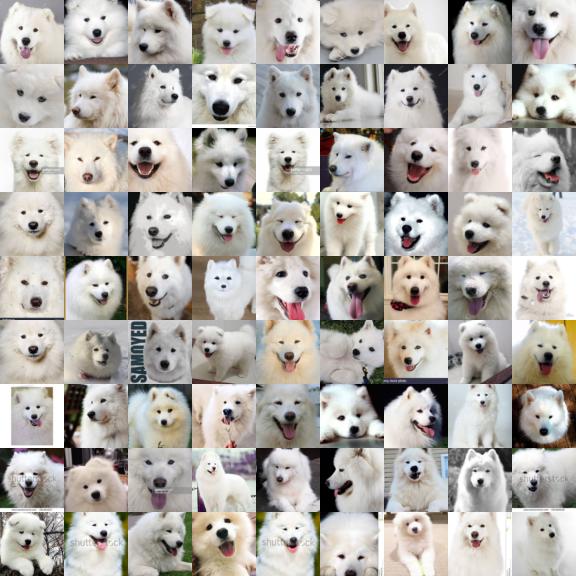}
\end{center}
  \caption{ Samples of the Zoo-Faces dataset  }
\label{fig:dataset-8}
\end{figure}

  \begin{figure}[h]
\begin{center}
  \includegraphics[width=0.5\linewidth]{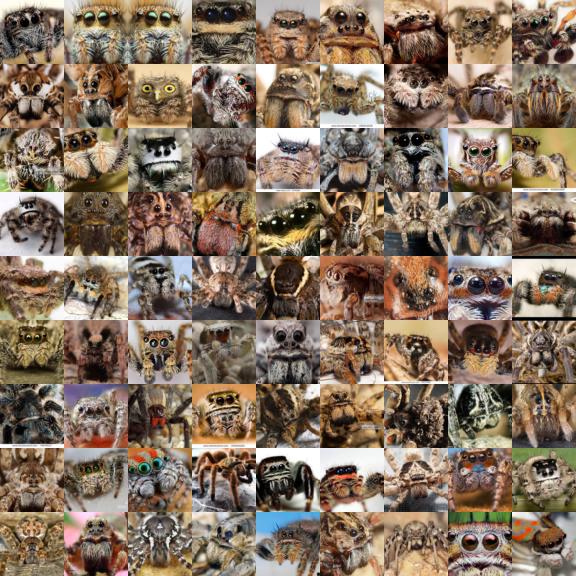}
\end{center}
  \caption{ Samples of the Zoo-Faces dataset   }
\label{fig:dataset-9}
\end{figure}

  \begin{figure}[h]
\begin{center}
  \includegraphics[width=0.5\linewidth]{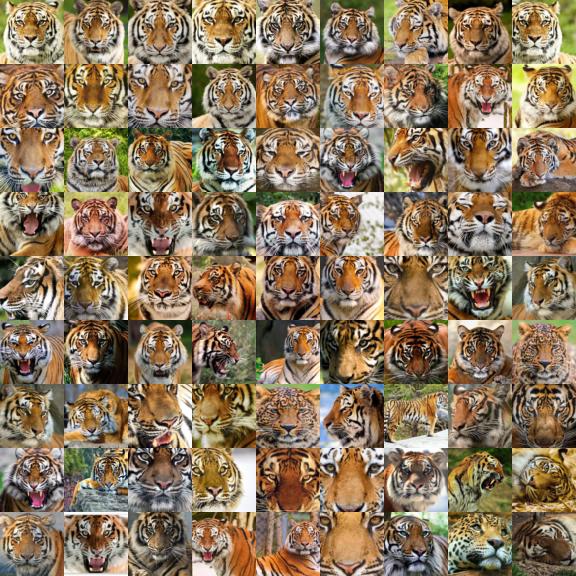}
\end{center}
  \caption{ Samples of the Zoo-Faces dataset   }
\label{fig:dataset-10}
\end{figure}

\clearpage

\section{Samples of the K-GAN, K-GAN + CycleGAN Models }
  \begin{figure}[h]
\begin{center}
  \includegraphics[width=0.5\linewidth]{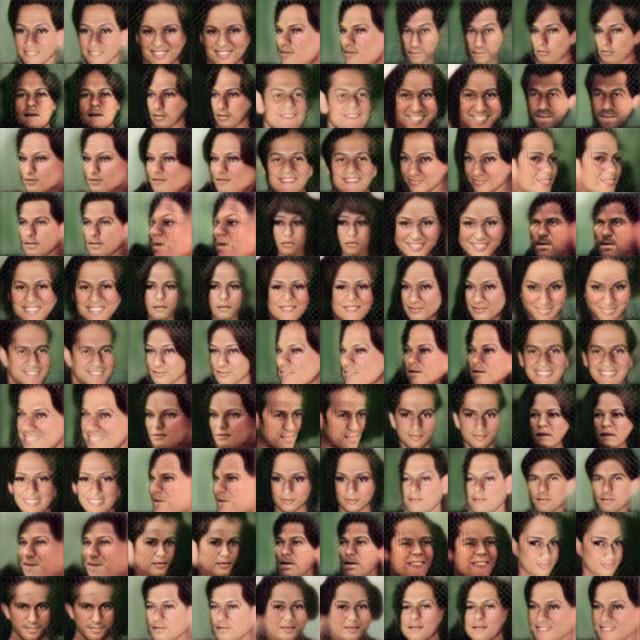}
\end{center}
  \caption{ Samples of K-GAN model trained on Apes and faces }
\label{K-apes}
\end{figure}


  \begin{figure}[h]
\begin{center}
  \includegraphics[width=0.5\linewidth]{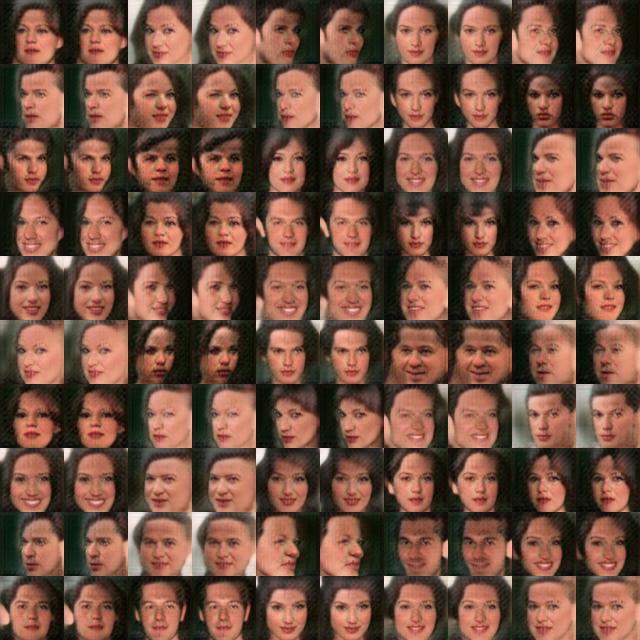}
\end{center}
  \caption{ Samples of K-GAN model trained on egyptian-cat and faces }
\label{k-bullmastiff}
\end{figure}



  \begin{figure}[h]
\begin{center}
  \includegraphics[width=0.5\linewidth]{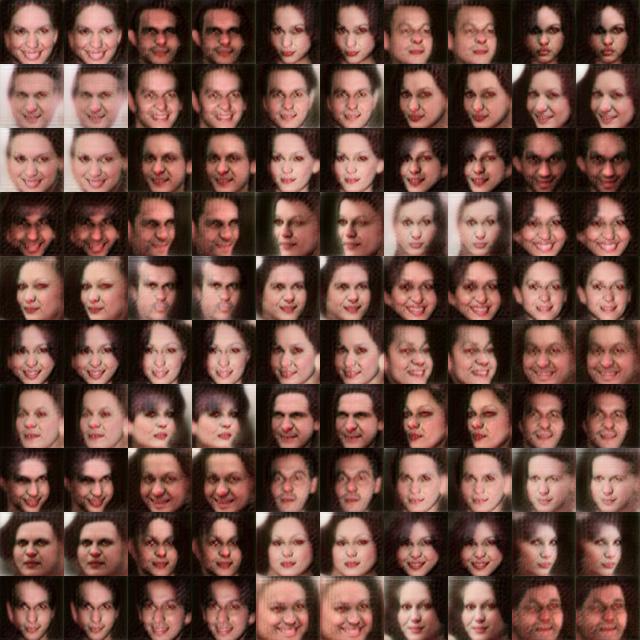}
\end{center}
\caption{ Samples of K-GAN model trained on chowchow and faces }
\label{k-chowchow}
\end{figure}





  \begin{figure}[h]
\begin{center}
  \includegraphics[width=0.5\linewidth]{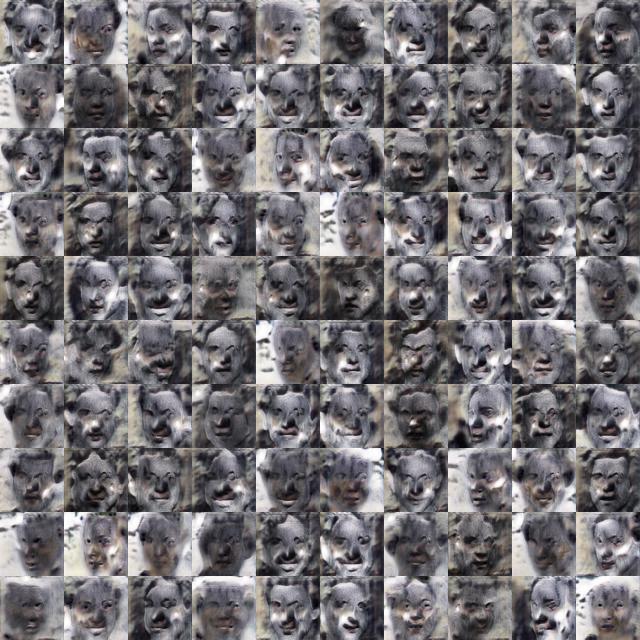}
\end{center}
\caption{Samples of full IAN model trained on koalas  and faces }
\label{c-hamsters}
\end{figure}

  \begin{figure}[h]
\begin{center}
  \includegraphics[width=0.5\linewidth]{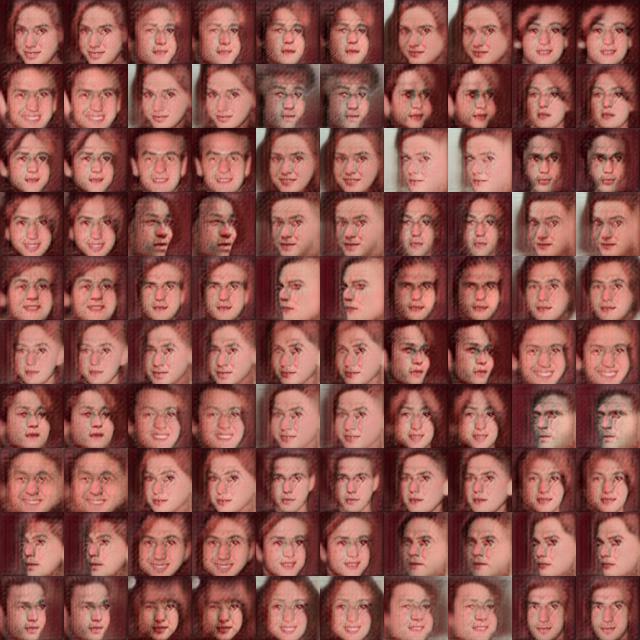}
\end{center}
\caption{ Samples of K-GAN model trained on lions  and faces }
\label{K-lions}
\end{figure}




  \begin{figure}[h]
\begin{center}
  \includegraphics[width=0.5\linewidth]{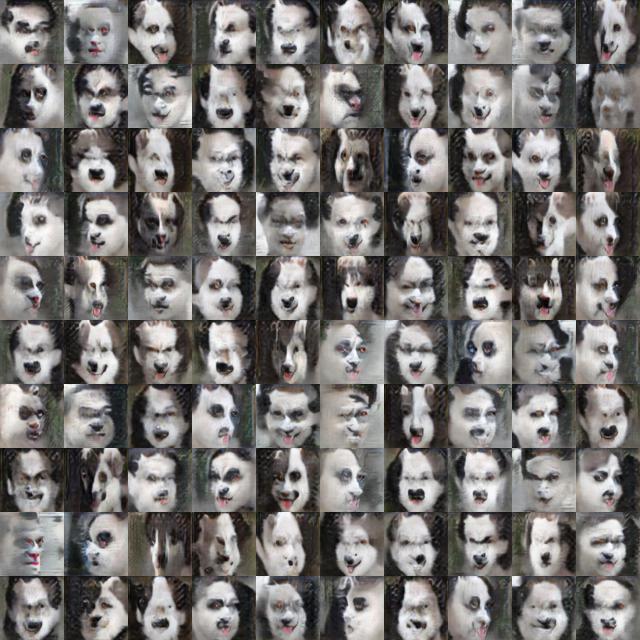}
\end{center}
\caption{ Samples of full IAN model trained on norwegian elkhoundand faces }
\label{c-norwegian-elkhound}
\end{figure}

  \begin{figure}[h]
\begin{center}
  \includegraphics[width=0.5\linewidth]{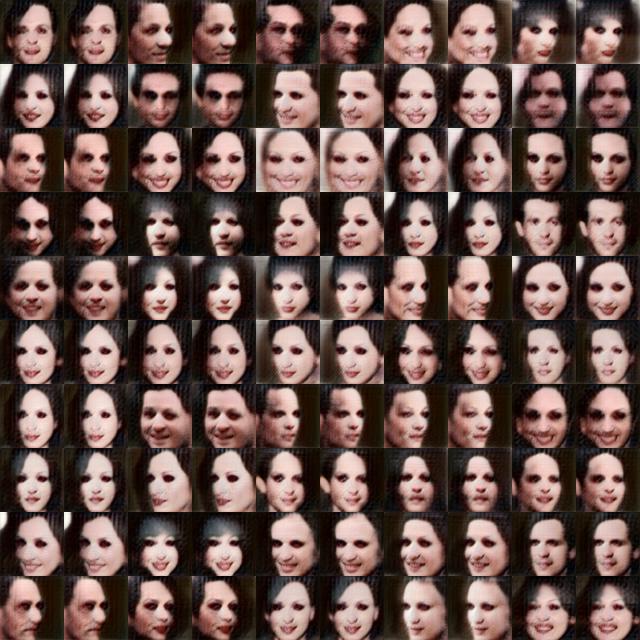}
\end{center}
\caption{ Samples of K-GAN model trained on pandas  and faces }
\label{K-pandas}
\end{figure}





  \begin{figure}[h]
\begin{center}
  \includegraphics[width=0.5\linewidth]{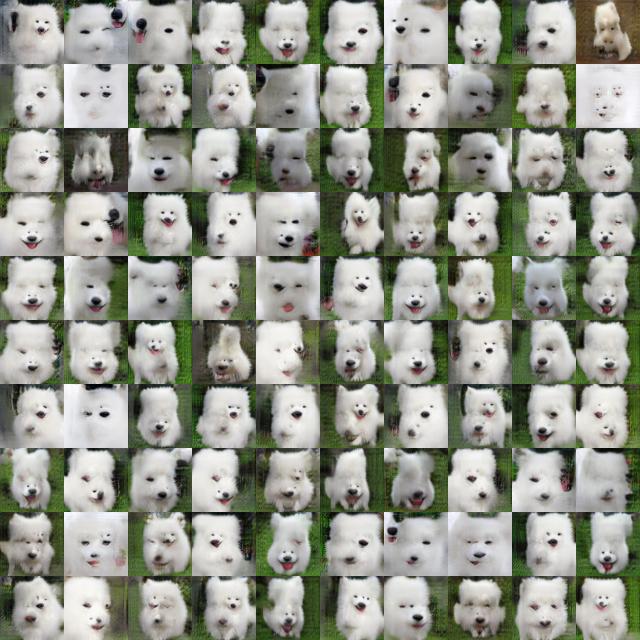}
\end{center}
\caption{ Samples of full IAN model trained on samoyed dog and faces }
\label{c-samoyed-dog}
\end{figure}

  \begin{figure}[h]
\begin{center}
  \includegraphics[width=0.5\linewidth]{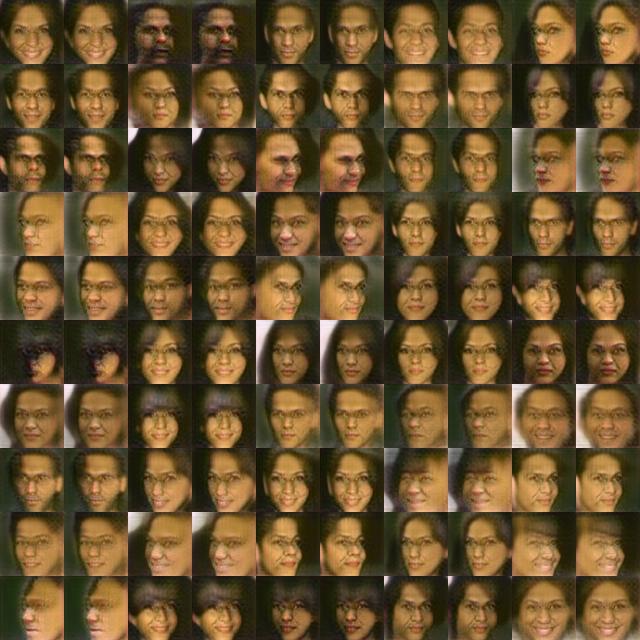}
\end{center}
\caption{ Samples of K-GAN model trained on spiders and faces }
\label{K-spiders}
\end{figure}
 



  \begin{figure}[h]
\begin{center}
  \includegraphics[width=0.5\linewidth]{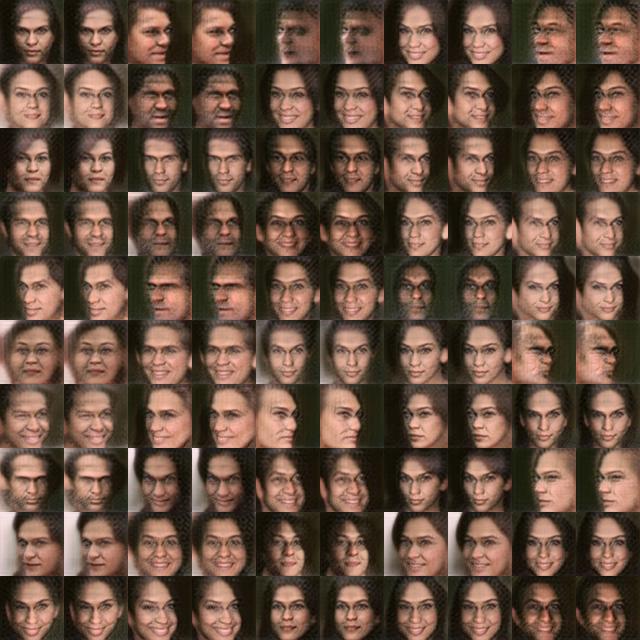}
\end{center}
\caption{ Samples of K-GAN model trained on tigers  and faces }
\label{K-tigers}
\end{figure}



\clearpage


\section{Confusion Matrices for Human classification experiment }

  \begin{figure}[h]
  \includegraphics[width=0.8\linewidth]{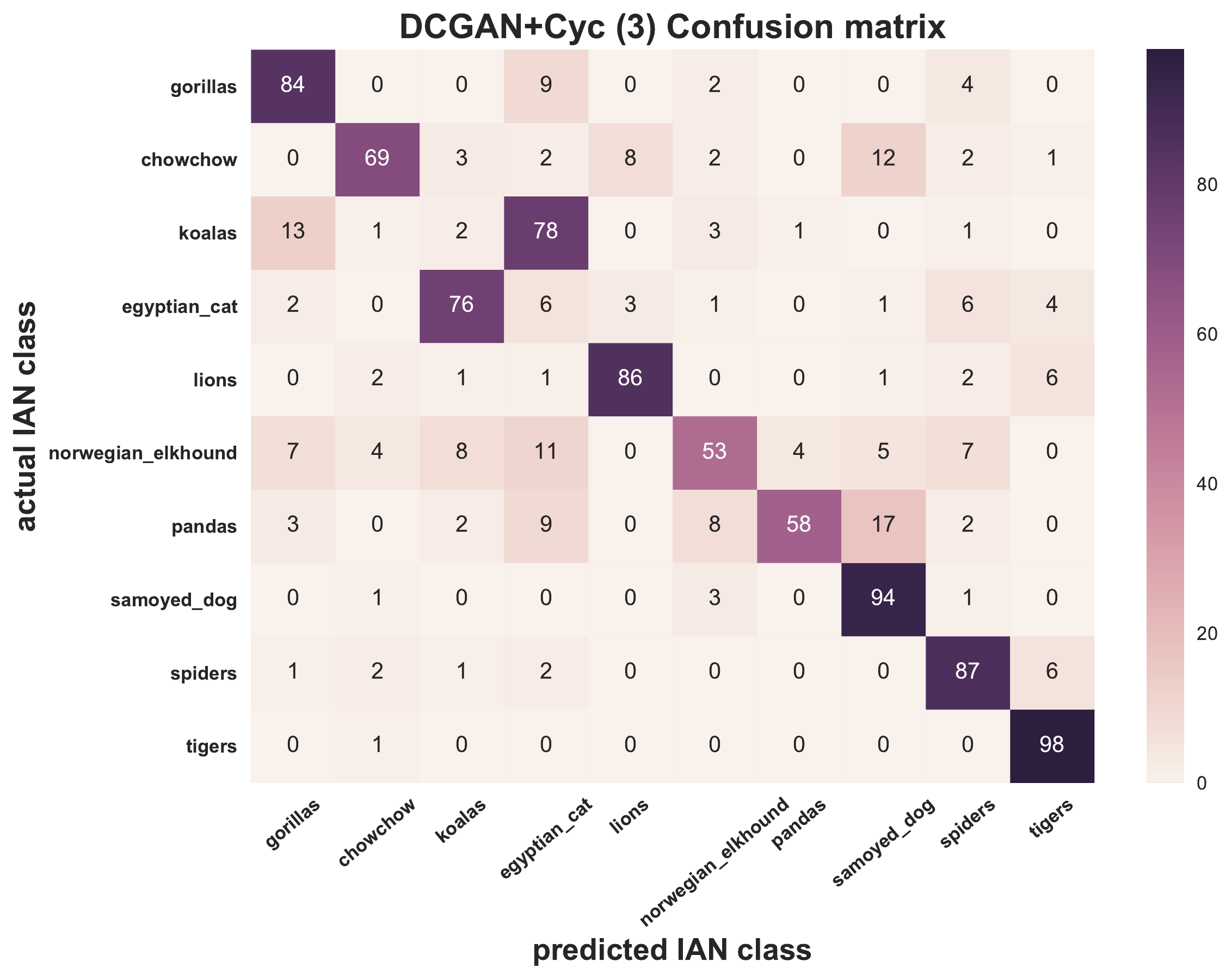}
\caption{ Confusion Matrix for DCGAN + CycleGAN }
\label{DCGAN-cyc}
\end{figure}

  \begin{figure}[h]
  \includegraphics[width=0.8\linewidth]{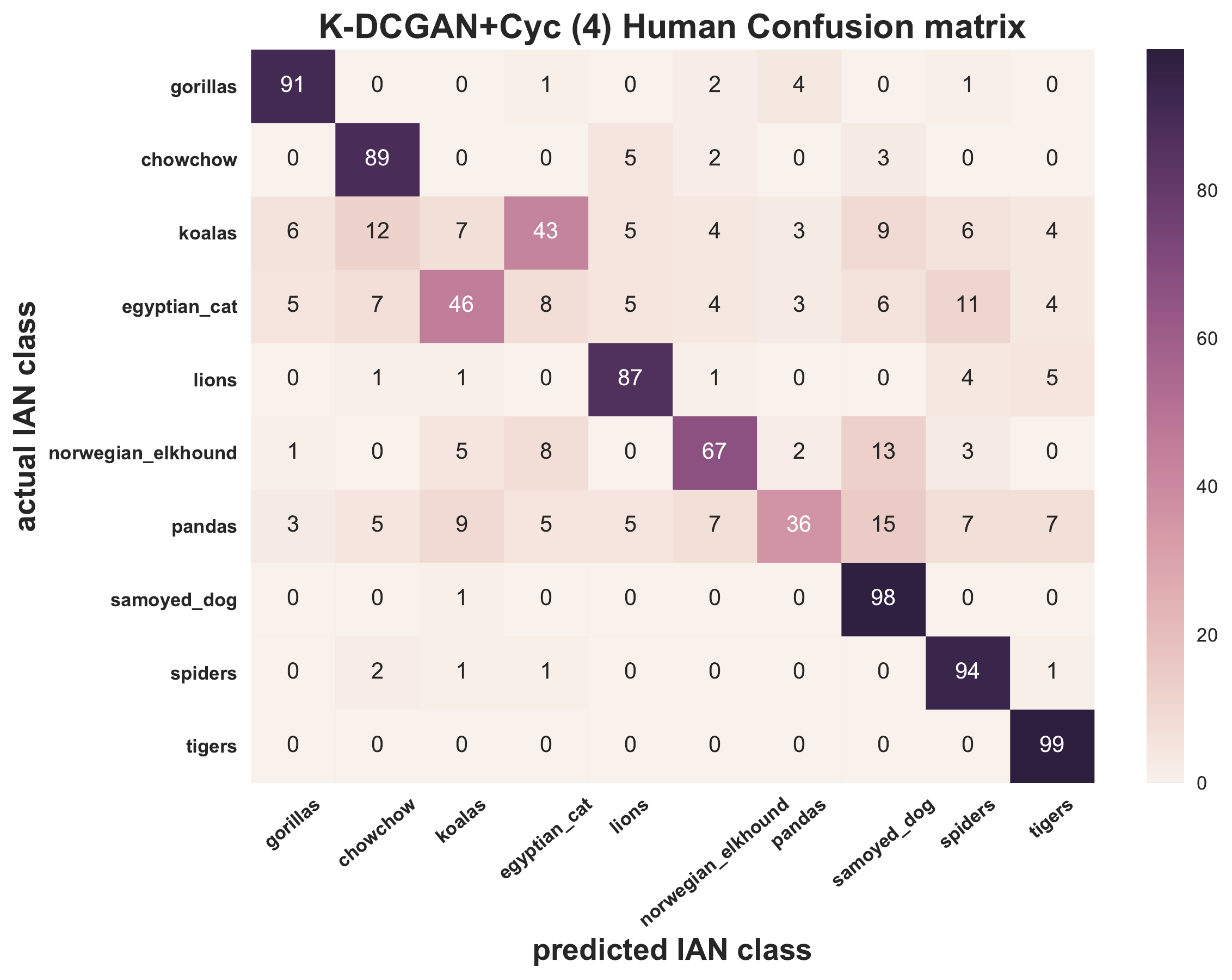}
\caption{ Confusion Matrix for K-DCGAN + CycleGAN }
\label{kDCGAN-cyc}
\end{figure}

  \begin{figure}[h]
  \includegraphics[width=0.8\linewidth]{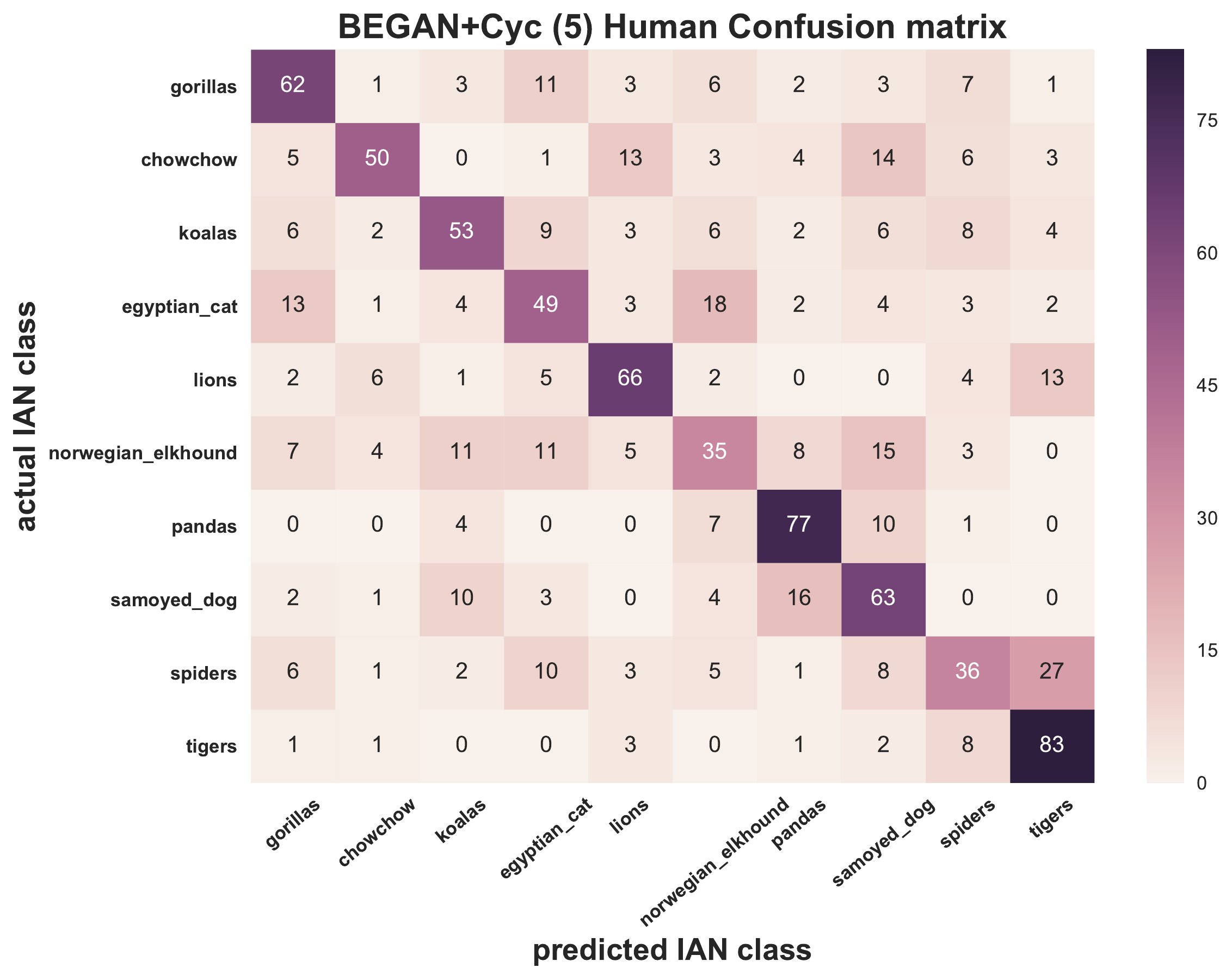}
\caption{ Confusion Matrix for BEGAN + CycleGAN }
\label{beGAN-cyc}
\end{figure}

  \begin{figure}[h]
  \includegraphics[width=0.8\linewidth]{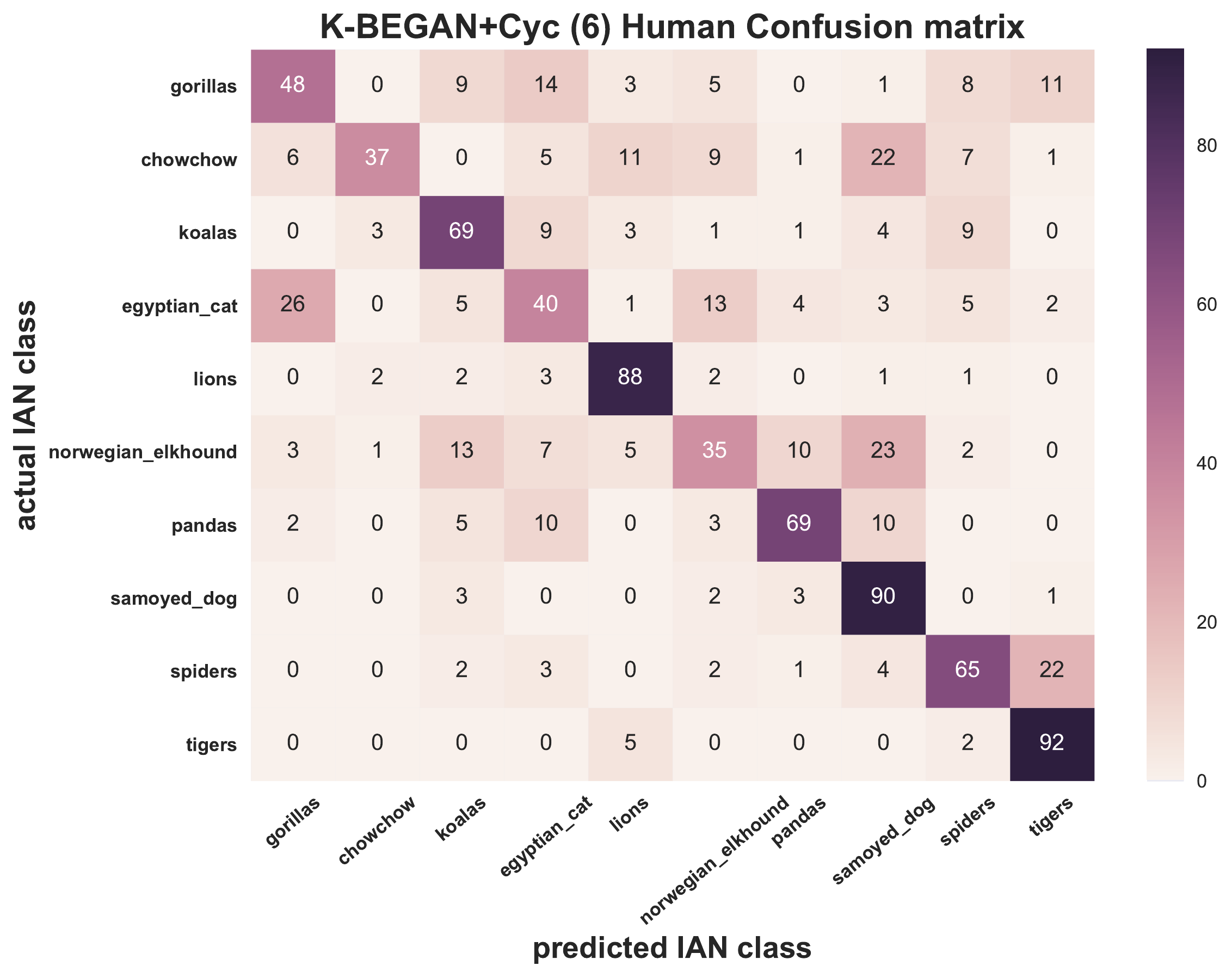}
\caption{ Confusion Matrix for K-BEGAN + CycleGAN }
\label{kbeGAN-cyc}
\end{figure}

\end{document}